\documentclass{article}

    \PassOptionsToPackage{numbers, compress}{natbib}

\usepackage[final]{neurips_2021}

\usepackage[utf8]{inputenc} 
\usepackage[T1]{fontenc}    
\usepackage{hyperref}       
\usepackage{url}            
\usepackage{booktabs}       
\usepackage{amsfonts}       
\usepackage{nicefrac}       
\usepackage{microtype}      
\usepackage{xcolor}         

\usepackage{graphicx}
\usepackage{subfigure}

\usepackage{algorithm}
\usepackage{algorithmic}

\usepackage{amsmath,amscd,amssymb,amsthm}
\usepackage{verbatim}
\usepackage{thmtools}
\usepackage{thm-restate}

\declaretheorem[name=Theorem]{Theorem}

\declaretheorem[name=Lemma, numberlike=Theorem]{Lemma}

\declaretheorem[sibling=Theorem]{Corollary}

\usepackage[framemethod=TikZ]{mdframed}
\mdfdefinestyle{MyFrame}{%
    linecolor=black,
    outerlinewidth=2pt,
    roundcorner=20pt,
    innerrightmargin=20pt,
    innerleftmargin=20pt,
    backgroundcolor=white}
\usepackage{hyperref}

\newcommand{\R}{\mathbb{R}}
\newcommand{\E}{\mathop{\mathbb{E}}}

\newcommand{\sign}{\text{sign}}

\newcommand{\bw}{\vec{w}}
\newcommand{\bx}{\vec{x}}
\newcommand{\bz}{\vec{z}}

\newcommand{\bm}{\vec{m}}
\newcommand{\bg}{\vec{g}}

\newcommand{\bgclip}{\vec{g}^{\text{clip}}}

\newcommand{\power}{\mathfrak{p}}

\newcommand{\one}{{\mathbf 1}}
\usepackage{graphicx}

\newcommand{\B}{\mathcal{B}}
\title{High-probability bounds for Non-Convex Stochastic Optimization with Heavy Tails}

\author{%
  Ashok Cutkosky\\
  Boston University\\
  \texttt{ashok@cutkosky.com} \\
   \And
   Harsh Mehta \\
   Google Research \\
   \texttt{harshm@google.com} \\
}

\begin{document}

\maketitle

\begin{abstract}
We consider non-convex stochastic optimization using first-order algorithms for which the gradient estimates may have heavy tails. We show that a combination of gradient clipping, momentum, and normalized gradient descent yields convergence to critical points in high-probability with best-known rates for smooth losses when the gradients only have bounded $\mathfrak{p}$th moments for some $\mathfrak{p}\in(1,2]$. We then consider the case of second-order smooth losses, which to our knowledge have not been studied in this setting, and again obtain high-probability bounds for any $\mathfrak{p}$. Moreover, our results hold for arbitrary smooth norms, in contrast to the typical SGD analysis which requires a Hilbert space norm. Further, we show that after a suitable ``burn-in'' period, the objective value will \emph{monotonically decrease} whenever the current iterate is not a critical point, which provides intuition behind the popular practice of learning rate ``warm-up'' and also yields a last-iterate guarantee.
\end{abstract}

\section{SGD and Heavy Tails}\label{sec:intro}

The standard algorithm for training large machine learning models today is stochastic gradient descent (SGD) and its myriad variants \citep{ghadimi2013stochastic, duchi10adagrad, kingma2014adam, you2017large, reddi2018on}. SGD enjoys many properties that help explain its empirical success: it can be implemented very easily, can efficiently process training data in a streaming manner, and even achieves the optimal convergence rate for finding critical points of smooth non-convex objectives \citep{arjevani2019lower}. In particular, SGD can be understood as optimizing the following stochastic optimization problem:
\begin{align*}
    \min_{\bw} F(\bw)= \E_z[f(\bw,z)]
\end{align*}
where $f(\bw,z)$ is some loss function taking model parameters $\bw$ and a random example $z$, where $z$ has some distribution $P_z$. Then SGD consists of the update:
\begin{align*}
    \bw_{t+1} = \bw_t - \eta \nabla f(\bw_t,z_t)
\end{align*}
where $z_1,\dots,z_T$ are i.i.d. random variables distributed according to $P_z$, and $\eta$ is some real number called the learning rate. Assuming that $F$ is $L$-smooth and that $\nabla f(\bw,z)$ satisfies $\E[\|\nabla f(\bw ,z)\|^2]\le G^2$ for some $G$ for all $\bw$, SGD satisfies \cite{ghadimi2013stochastic}:
\begin{align*}
    \frac{1}{T}\E\left[\sum_{t=1}^T \|\nabla F(\bw_t)\|^2\right]&\le O\left(\frac{L\Delta}{T} + \frac{G^2}{\sqrt{T}}\right)
\end{align*}
where $\Delta= F(\bw_1)-\inf F(\bw)$. This implies that SGD takes $O(\epsilon^{-4})$ iterations to find an $\epsilon$-critical point, which is a point $\bw$ such that $\|\nabla F(\bw)\|\le \epsilon$ \citep{ghadimi2013stochastic}. When the stochastic gradients are further assumed to have light sub-gaussian tails, \citep{li2020high} shows that SGD with momentum can achieve this same result in high probability as well.

However, recent work \citep{simcsekli2019heavy,simsekli2019tail,zhang2019gradient,zhang2020improved,zhang2020adaptive} suggests that the assumptions of bounded variance or light tails may be too optimistic in practical deep learning problems. For example, \citet{zhang2020adaptive} provides empirical evidence that large NLP models based on attention and transformers \citep{vaswani2017attention,devlin-etal-2019-bert} have \emph{heavy-tailed} gradients. In particular, the variance is extremely large (perhaps even essentially infinite for practical purposes), but the $\power th$ moment is bounded for some $\power\in (1,2]$. This setting is significantly more complicated, and SGD's convergence properties are difficult to characterize even when assuming slightly stronger tail bounds \cite{scaman2020robustness} or various versions of convexity \cite{wang2021convergence}. 

A standard tool when working with heavy-tailed random variables is truncation, or \emph{clipping}. Succinctly, if $X$ is a random variable such that $\E[\|X\|^2]=\infty$, $X$ itself may be poorly behaved due to the heavy tail, but a truncated value $X^{\text{clip}}= \frac{X}{|X|}\min(|X|, \tau)$ will be much more benign so long as $\tau$ is chosen in an appropriate manner \citep{bubeck2013bandits, lugosi2019mean}. Fortuitously, gradient clipping has long been a standard technique to improve training of deep neural networks \citep{pascanu2012understanding}, and recent analysis \citep{zhang2020adaptive, gorbunov2020stochastic} suggests that clipping allows SGD to deal with heavy-tailed stochastic gradients. Specifically, \citep{zhang2020adaptive} shows that SGD combined with clipping can find an $\epsilon$-critical point in expectation in $O\left(\epsilon^{-\frac{3\power-2}{\power-1}}\right)$ iterations.

We propose to improve upon this backdrop in three important ways. First, the prior analysis of \citep{zhang2020adaptive} for non-convex convergence of SGD under heavy tails only provides an \emph{in-expectation} analysis. In-expectation results are unsatisfying because the quantity measured for convergence (i.e. gradient norm or function value) is itself a random variable, and so might \emph{also} have a heavy tail. In this case individual training runs could be quite poor. Since training modern networks is quite expensive \citep{JMLR:v21:20-074:T5-paper, brown2020language}, it is important to be confident that each and every training run will produce a high-quality result. We provide a variant of the normalized SGD with momentum algorithm of \citep{cutkosky2020momentum} that incorporates gradient clipping and show that the method finds an $\epsilon$-critical point in $\tilde O\left(\epsilon^{-\frac{3\power-2}{\power-1}}\right)$ iterations with high probability.

Second, the vast majority of first-order optimization methods for stochastic non-convex optimization consider exclusively the standard $L_2$ or Hilbert-space norm. Such analyses can take advantage of the fact that $\E[\langle \nabla F(x), \bg\rangle]=\|\nabla F(x)\|^2$ when $\bg$ is an unbiased estimate of $\nabla F(x)$. In contrast, our algorithms find critical points when the gradient is measured with any smooth norm (i.e. in a Banach space). In this case, the gradient estimate $\bg$ (which is a dual vector) must be converted into an appropriate primal vector. This operation is non-linear in general and so may add some bias. It turns out the key to operating with different norms is the popular practice of augmenting SGD iterates with momentum. When using momentum-based updates we do not rely on the same unbiasedness argument, and so the techniques can be extended to more general norms. This provides more concrete evidence of the theoretical advantages of momentum.

Third, recent theoretical advances in non-convex optimization have produced a number of new algorithms that avoid the lower-bounds of \citep{arjevani2019lower} by assuming extra structure on the objective $F$, such as \emph{second-order smoothness}. In this case, \citep{tripuraneni2018stochastic,allen2018natasha,fang2019sharp,cutkosky2020momentum,arjevani2020second} provide algorithms that achieve faster convergence rates. In particular, the algorithms of \citep{fang2019sharp,cutkosky2020momentum} can find an $\epsilon$-critical point in $O(\epsilon^{3.5})$ iterations using a first-order stochastic oracle, which is the optimal rate \citep{arjevani2020second}. To our knowledge, no analysis even in expectation exists for second-order smooth losses under heavy tails. We fill this gap, providing an algorithm that identifies an $\epsilon$-critical point in $\tilde O\left(\epsilon^{-\frac{5\power-3}{2\power-2}}\right)$ iterations with high probability. Further, in both smooth and second-order smooth cases we provide a convergence guarantee for the \emph{last iterate} of the optimization algorithm: with high probability, the objective value for the final iterate will be smaller than the objective value at some approximate critical point.

Finally, we discuss an intriguing connection to the popular \emph{warm-up} learning rate schedule \citep{He2015DeepRL,goyal2017accurate,devlin-etal-2019-bert}, in which the learning rate starts small and increases over the first iterations of training. We show that for normalized SGD with momentum, after a brief ``burn-in'' period, the objective value decreases monotonically until an approximate critical point is reached. This provides an alternative motivation for warm-up: it would be reasonable to keep the learning rate small during the ``burn-in'' period. There is a long line of empirical research in understanding the need for warm-up. Warm-up has been shown to be connected to the use of layer-normalization, and that improved use of layer-norm can enable warm-up free training of transformer models \citep{Chen_2018_nmt_warmup, Wang_2019,Nguyen2019TransformersWT}. \citet{pmlr-v119-huang20f} further build on these insights and proposes a weight initialization scheme which removes the need for both layer normalization and warm-up while training transformer models. Finally, \citet{Liu2020On} demonstrate that the variance of the learning rate in optimizers like Adam is abnormally high in the early stages of the optimization process and further hypothesize that warm-up aids by reducing the variance in that stage.  Our results provide a different view from an optimization perspective.

\textbf{Paper Organization:} in Section \ref{sec:assumptions} we provide our concrete assumptions and setup. In Sections \ref{sec:normalized} and \ref{sec:igt}, we provide our algorithms and high-probability convergence bounds. In Section \ref{sec:experiments}, we provide some empirical observations on the effect of the burn-in phase in our analysis.

\subsection{Setup and assumptions}\label{sec:assumptions}
We assume that the objective $F(\bw)=\E[f(\bw,z)]$ is a Frechet-differentiable function mapping a Banach space $\B$ to the reals $\R$. Let $\|\cdot\|$ be the norm in $\B$ and $\|\cdot\|_\star$ be the dual norm in the dual space $\B^\star$. To preserve analogy and notation from the Hilbert space setting, we will write $\langle v,w\rangle$ to indicate the application of a dual vector $v\in \B^\star$ to a vector $w\in \B$. We assume that $\|\cdot\|_\star^p$ is Frechet differentiable and satisfies for some $p,C$ for all $x,y\in \B$:
\begin{align}
    \|x+y\|_\star^p\le \|x\|_\star^p+\langle \nabla \|x\|_\star^p,y\rangle + C\|y\|_\star^p \label{eqn:banach}
\end{align}
Such a Banach space is called $(p,C)$-smooth. Note that if $\|\cdot\|$ is the standard $p$ norm in $\R^n$ for $p\in(1,2]$, then $\|\cdot\|$ is the corresponding $q$ norm, and the dual space is $(2,\frac{1}{p-1})$-smooth. In the common case that $\B$ is a Hilbert space (e.g. $\R^n$ with the euclidean norm), then $\B=\B^\star$, $\|\cdot\|=\|\cdot\|_\star$ and $p=2$ and $C=1$. In this paper, we will exclusively consider the case $p=2$ and $C\ge 1$. Finally, given any $v\in \B^\star$, we will write $d(v)\in \B$ to indicate a unit vector satisfying both $\|d(v)\|=1$ and $\langle v,d(v)\rangle =\|v\|_\star$, which we assume exists for all $v$. For more background on Banach spaces, see Appendix \ref{sec:smoothbanach}.

We assume $F$ is $L$-smooth, which means that $\|\nabla F(\bw) - \nabla F(\bx)\|_\star\le L\|\bw-\bz\|_\star$. In Section \ref{sec:igt}, we will also assume that $F$ is $\rho$-second-order smooth, which means that $\|(\nabla^2 F(\bw) - \nabla^2F(\bx))v\|_\star\le \rho \|\bw-\bx\|\|v\|$ for all $\bw,\bx,v\in \B$. We assume that $\nabla f(\bw,z)$ satisfies both $\E[\|\nabla f(\bw,z)-\nabla F(\bw)\|_\star^\power]\le G^\power$ and $\E[\|\nabla f(\bw ,z)\|_\star^\power]\le G^\power$ for some $G$ and some $\power\in (1,2]$. We will design algorithms such that after $T$ stochastic gradient evaluations, we can output a point $\bw$ such that $\|\nabla F(\bw)\|_\star$ is small with high probability.

\section{Normalized SGD with Momentum}\label{sec:normalized}
In this section, we describe our algorithms for finding critical points with high probability in non-convex objectives. Our analysis makes use of gradient clipping, momentum, and normalized updates. Clipping is a standard technique for mitigating the effect of heavy tails, essentially by throwing out outlier data. Momentum is useful as it intuitively averages together many clipped gradients, resulting in a gradient estimate that concentrates about its mean with high probability. Normalizing the updates allows for significantly simplified analyses because the bias terms introduced by the moving average in momentum are now very precisely controlled by the learning rate, as originally analyzed for Hilbert spaces in \citep{cutkosky2020momentum}. In order to analyze this algorithm, we need the following critical Lemma, which characterizes the effect of taking a normalized SGD step (proved in appendix \ref{sec:smoothbanach}).

\begin{restatable}{Lemma}{lemonestep}\label{lem:onestep}
Suppose $F:\B\to \R$ be a Frechet-differentiable $L$-Smooth function from a Banach space $\B$ to the reals. Let $w\in B$. Let $g^\star\in \B^\star$ and let $g\in \B$ be a unit-vector satisfying $\langle g^\star, g\rangle = \|g\|_\star$. Define $w' = w-\eta g$. Define $\epsilon = g^\star-\nabla F(w)$. Then:
\begin{align*}
    F(w') \le F(w) - \eta \|\nabla F(w)\|_\star + 2\eta \|\epsilon\|_\star + \frac{L\eta^2}{2}
\end{align*}
\end{restatable}
\begin{algorithm}
   \caption{Normalized SGD with Clipping and Momentum}
   \label{alg:nsgdclip}
   \begin{algorithmic}
      \STATE{\bfseries Input: } Initial Point $\bw_1$, learning rate $\eta$, momentum parameter $\beta$, clipping parameter $\tau$, time horizon $T$:
      \STATE Set $\bm_0 = 0$.
      \FOR{$t=1\dots T$}
      \STATE Sample $z_t\sim P_z$.
      \STATE Set $\bgclip_t = \frac{\nabla f(\bw_t, z_t)}{\|\nabla f(\bw_t, z_t)\|_\star}\min(\tau, \|\nabla f(\bw_1,z_t)\|_\star)$.
      \STATE Set $\bm_t = \beta \bm_{t-1}+ (1-\beta) \bgclip_t$.
      \STATE Set $\bw_{t+1} = \bw_t  - \eta d(\bm_t)$.
      \ENDFOR
   \end{algorithmic}
\end{algorithm}

\begin{restatable}{Theorem}{thmnsgdclip}\label{thm:nsgdclip}
Suppose $\E_z[\|\nabla f(\bw,z)\|_\star^\power]\le G^\power$ for all $\bx$ for some $G$. Suppose $F$ is $L$-smooth. Set $\beta=1-\alpha$, $\alpha = \frac{b}{T^{\frac{\power}{3\power-2}}}$ and $\eta = \frac{s}{T^{\frac{2\power -1}{3\power-2}}}$ for arbitrary constant $b$ and $s$ satisfying $\alpha \le 1$ and set $\tau=G/\alpha^{1/\power}$. Then with probability at least $1-\delta$:
\begin{align*}
    \frac{1}{T}\sum_{t=1}^T \|\nabla F(\bw_t)\|_\star&\le  O\left(\frac{\log(T/\delta)}{T^{\frac{\power-1}{3\power-2}}}\right)
\end{align*}
where the big-Oh hides constant that depend on $L$, $G$ $b$, $s$, $C$ and $F(\bw_1)-F(\bw_{T+1})$, but not $T$ or $\delta$.
\end{restatable}
This Theorem provides an analog of the standard in-expectation result for non-convex optimization, but now the result holds with high-probability. Note that this matches the in-expectation convergence rates from \citet{zhang2020adaptive} up to logarithmic factors. The next Theorem takes this theme one step further: it shows that after a brief ``burn-in'' period, the decrease in objective value will in fact be \emph{monotonic} with high probability:
\begin{restatable}{Theorem}{thmnsgdclipburnin}\label{thm:nsgdclipburnin}
Under the assumptions of Theorem \ref{thm:nsgdclip}, define the constants:
\begin{align*}
K&= 10C\max(1,\log(3T/\delta)) + 4C^{1/2}\sqrt{\max(1,\log(3T/\delta))} + 1\\
    Z&= \frac{sL}{b}+GKb^{\frac{\power-1}{\power}}
\end{align*}
Then, with probability at least $1-\delta$, for all $t$ satisfying:
\begin{align*}
t\ge \mathcal{T}=\frac{T^\frac{\power}{3\power-2}}{b}\left(\frac{\power-1}{3\power-2}\log(T) +\log(G)- \log(Z)\right)
\end{align*}
we have
\begin{align*}
    \|\bm_t-\nabla F(\bw_t)\|_\star&\le\frac{2Z}{T^{\frac{\power-1}{3\power-2}}}
\end{align*}
Further, so long as
\begin{align}
\|\bm_t\|_\star\ge 2\left(\frac{6 Z}{T^{\frac{\power-1}{3\power-2}}} + \frac{Ls}{2T^{\frac{2\power-1}{3\power-2}}}\right)\label{eqn:mcondition}
\end{align}
we have $F(\bw_{t+1})< F(\bw_t)-\frac{\eta}{2}\|\bm_t\|_\star$. Moreover, whenever (\ref{eqn:mcondition}) is not satisfied for $t\ge \mathcal{T}$, we must have 
\begin{align*}
    \|\nabla F(\bw_t)\|_\star\le \frac{14 Z}{T^{\frac{\power-1}{3\power-2}}} + \frac{Ls}{T^{\frac{2\power-1}{3\power-2}}}
\end{align*}
Finally, if $t$ is the iteration with smallest value of $\|\bm_t\|_\star$ such that $t\ge \mathcal{T}$, we have:
\begin{align*}
\|\nabla F(\bw_t)\|_\star \le O\left(\frac{\log(T/\delta)}{T^{\frac{\power-1}{3\power-2}}}\right)
\end{align*}
\end{restatable}

Theorem \ref{thm:nsgdclipburnin} shows that after a small burn-in period $\mathcal{T}=o(T)$, $\bm_t$ is very likely to be an extremely good estimate of the true gradient $\nabla F(\bx_t)$. Moreover, we can empirically detect this event by monitoring the norm of the exponentially weighted moving average $\bm$. This burn-in period is especially interesting given the success of learning rate schedules that incorporate a ``warm-up'' component, in which the learning rate actually increases during the first iterations. Our result seems to hint at a mechanism for this schedule: by keeping the learning rate small during the burn-in period, we can limit any increase in the objective value until we reach the stage at which it begins to decrease steadily.

Further, Theorem \ref{thm:nsgdclipburnin} provides a very concrete algorithm to find critical points: after the burn-in period of $\tilde O(T^{\frac{\power}{3\power-2}})$ iterates, we can simply return the point $\bx_t$ with the smallest value of $\|\bm_t\|$. Intuitively, this works because the objective value can only increase by a total of $O(T^{\frac{\power}{3\power-2}}\eta)=O(T^{\frac{1-\power}{3\power-2}})=O(1)$ during the burn-in period so even without a learning rate ``warm-up'', we should not expect the burn-in period to have too significant an effect on the objective value. After the burn-in period, $\bm_t$ is always very close to $\nabla F(\bw_t)$, so we make steady progress until a critical point is encountered.

\begin{proof}[Proof of Theorem \ref{thm:nsgdclip}]
Define $\hat \epsilon_t = \bm_t - \nabla F(\bw_t)$. and $\epsilon_t = \bgclip_t - \nabla F(\bw_t)$. Also, define $S(a,b)=\nabla F(a)-\nabla F(b)$. By smoothness, we have $\|S(\bw_t,\bw_{t+1})\|_\star\le L\|\bw_t-\bw_{t+1}\|_\star$

From the definition of $\bm_{t+1}$, we have the following recursive formulation for any $t\ge 1$:
\begin{align}
    \bm_{t+1} &= (1-\alpha) (\nabla F(\bw_t) + \hat \epsilon_t) + \alpha\bg^{\text{ clip}}_{t+1}\nonumber\\
    &=\nabla F(\bw_{t+1}) + (1-\alpha)(S(\bw_t,\bw_{t+1})+\hat \epsilon_t) + \alpha\epsilon_{t+1}\nonumber\\
    \hat \epsilon_{t+1}&= (1-\alpha)S(\bw_t,\bw_{t+1})+(1-\alpha)\hat \epsilon_t + \alpha \epsilon_{t+1}\label{eqn:epsrecursionsmooth}
\end{align}
Further, by defining $\bm_0=0$ and $\bw_0=\bw_1$ so that $\hat \epsilon_0 = -\nabla F(\bw_1)$, we have that (\ref{eqn:epsrecursionsmooth}) holds for $t=0$ as well.
Now, we unravel the recursion for all iterations:
\begin{align*}
    \hat \epsilon_{t+1}&=(1-\alpha)^{t+1}\hat \epsilon_0 +\alpha\sum_{t'=0}^{t} (1-\alpha)^{t'} \epsilon_{t+1-t'}+ (1-\alpha)\sum_{t'=0}^{t}(1-\alpha)^{t'} S(\bw_{t-t'},\bw_{t+1-t'})
\end{align*}
Next, take the magnitude of both sides and use triangle inequality (using the fact that $\|S(\bw_t,\bw_{t+1})\|_\star\le L\eta$):
\begin{align*}
    \|\hat \epsilon_{t+1}\|_\star&\le (1-\alpha)^{t+1}\|\hat\epsilon_0\|_\star +\alpha\left\|\sum_{t'=0}^{t} (1-\alpha)^{t'} \epsilon_{t+1-t'}\right\|_\star+ (1-\alpha)\eta L \sum_{t'=0}^{t}(1-\alpha)^{t'}\\
    &\le (1-\alpha)^{t+1} G + \frac{(1-\alpha)\eta L}{\alpha}+\alpha\left\|\sum_{t'=0}^{t} (1-\alpha)^{t'} \epsilon_{t+1-t'}\right\|_\star
\end{align*}
where in the second line we have used $\bm_0=0$ and $\|\nabla F(\bw_0)\|_\star\le G$, which follows since $\|\nabla F(\bw_0)\|_\star\le \E_z[\|\nabla f(\bw_0,z)\|_\star^\power]^{1/\power}=G$.

Now, we invoke Lemma \ref{thm:thresholdsimplebanach}, a concentration inequality in Banach space (see appendix \ref{sec:concentration}). This implies that with probability at least $1-\delta/T$:
\begin{align*}
    \left\|\sum_{t'=0}^{t} (1-\alpha)^{t'} \epsilon_{t+1-t'}\right\|_\star&\le  10C\tau\max(1,\log(3T/\delta))+\sum_{t'=0}^t \frac{(1-\alpha)^{t'}G^\power}{\tau^{\power-1}}\\
    &\quad\quad+4\left(C\sum_{t'=0}^t(1-\alpha)^{2t'}G^\power \tau^{2-\power}\right)^{1/2}\sqrt{\max(1,\log(3T/\delta))}\\
    &\le 10C\tau\max(1,\log(3T/\delta))+\frac{G^\power}{\alpha \tau^{\power-1}} +\frac{4\left(CG^\power \tau^{2 -\power}\right)^{1/2}\sqrt{\max(1,\log(3T/\delta))}}{\alpha^{1/2}}
\end{align*}

Therefore, with probability at least $1-\delta/T$:
\begin{align}
    \|\hat \epsilon_{t+1}\|_\star&\le (1-\alpha)^{t+1} G + \frac{(1-\alpha)\eta L}{\alpha} + 10C\alpha\tau\max(1,\log(3T/\delta))\nonumber\\
    &\quad+ 4\left( \alpha CG^\power\tau^{2-\power}\right)^{1/2}\sqrt{\max(1,\log(3T/\delta))}+\frac{G^\power}{ \tau^{\power-1}}\nonumber\\
    \intertext{
Now, with $\tau=\frac{G}{\alpha^{1/\power}}$, and $D_\delta=\max(1,\log(3T/\delta))$ this becomes:}
&\le (1-\alpha)^{t+1} G + \frac{(1-\alpha)\eta L}{\alpha} +G\alpha^{1-1/\power}\left[10 CD_\delta + 4C^{1/p}\sqrt{D_\delta} + 1\right]\nonumber\\
    &=(1-\alpha)^{t+1}G+ \frac{(1-\alpha)\eta L}{\alpha} + G\alpha^{1-1/\power}K\label{eqn:hatepsbound}
\end{align}
where we define the constant $K=10CD_\delta + 4C^{1/2}\sqrt{D_\delta} + 1$.

Now, we have from Lemma \ref{lem:onestep}:
\begin{align*}
   \sum_{t=1}^T \|\nabla F(\bw_t)\|_\star&\le \frac{F(\bw_1)-F(\bw_T)}{\eta } + \frac{LT\eta}{2}+2\sum_{t=1}^T \|\hat\epsilon_t\|_\star
    \intertext{so with probability at least $1-\delta$:}
    &\le \frac{F(\bw_1)-F(\bw_T)}{\eta } + \frac{LT\eta}{2}+2\sum_{t=1}^T(1-\alpha)^tG+ \frac{(1-\alpha)\eta L}{\alpha} + G\alpha^{1-1/\power}K\\
    &\le \frac{F(\bw_1)-F(\bw_T)}{\eta } + \frac{LT\eta}{2}+\frac{2G(1-\alpha)}{\alpha}+ \frac{2(1-\alpha)\eta T L}{\alpha} + 2GT\alpha^{1-1/\power}K
\end{align*}
Now set $\alpha = \frac{b}{T^{\frac{\power}{3\power-2}}}$ and $\eta = \frac{s}{T^{\frac{2\power -1}{3\power-2}}}$. This yields:
\begin{align*}
    \frac{1}{T}\sum_{t=1}^T \|\nabla F(\bw_t)\|&\le  O\left( T^{\frac{1-\power}{3\power-2}}\log(T/\delta)\right)
\end{align*}
\end{proof}

The proof of Theorem \ref{thm:nsgdclipburnin} follows essentially the same idea as that of Theorem \ref{thm:nsgdclip}. However, this time we use the explicit bound on $\|\hat\epsilon_t\|_\star$ and observe that after the burn-in period, the contribution from $\|\epsilon_1\|_\star$ has decreased exponentially to be insignificant, and so $\bm_t$ will be a very high-quality estimate of $\nabla F(\bx_t)$ at every single iteration.

\begin{proof}[Proof of Theorem \ref{thm:nsgdclipburnin}]
Use the settings $\eta=\frac{s}{T^{\frac{2\power-1}{3\power-2}}}$ and $\alpha = \frac{b}{T^{\frac{\power}{3\power-2}}}$ and notice that from equation (\ref{eqn:hatepsbound}) that with probability $1-\delta$, for all $t$:
\begin{align*}
    \|\hat \epsilon_{t+1}\|_\star&\le (1-\alpha)^t G + \frac{sL}{bT^{\frac{\power-1}{3\power-2}}}+\frac{GKb^{\frac{\power-1}{\power}}}{T^{\frac{\power-1}{3\power-2}}}=(1-\alpha)^t G + \frac{Z}{T^{\frac{\power-1}{3\power-2}}}
\end{align*}
Further, we have
\begin{align*}
    \log(1-\alpha)&\le -\alpha\\
    \log\left(G(1-\alpha)^t\right) &\le -t\alpha +\log(G)
\end{align*}
Therefore, if $t\ge \mathcal{T}$, then we have with probability at least $1-\delta$:
\begin{align*}
\|\hat \epsilon_{t+1}\|_\star&\le \frac{2Z}{T^{\frac{\power-1}{3\power-2}}}
\end{align*}
This implies the bound on $\|\nabla F(\bw_t)\|_\star$ when $\|\bm_t\|_\star$ is smaller than the threshold described in the Theorem statement. Next, again from Lemma \ref{lem:onestep},
\begin{align*}
    F(\bw_{t+1})&\le F(\bw_t) - \eta\|\nabla F(\bw_t)\|_\star + 2\eta\|\hat\epsilon_t\|_\star + \frac{L\eta^2}{2}\\
    &\le F(\bw_t) - \eta\|\bm_t\|_\star + 3\eta\|\hat\epsilon_t\|_\star + \frac{L\eta^2}{2}\\
    &\le F(\bw_t) - \eta\|\bm_t\|_\star +\eta\left(\frac{6Z }{T^{\frac{\power-1}{3\power-2}}} + \frac{Ls}{2T^{\frac{2\power-1}{3\power-2}}}\right)
\end{align*}
from which we can conclude that if $\|\bm_t\|_\star\ge 2\left(\frac{6Z }{T^{\frac{\power-1}{3\power-2}}} + \frac{Ls}{2T^{\frac{2\power-1}{3\power-2}}}\right)$, $F(\bw_{t+1})\le F(\bw_t) - \frac{\eta}{2}\|\bm_t\|_\star$.

Now, for the final part of the Theorem, observe that since $F$ is $G$-Lipschitz, we must have that 
\begin{align*}
    F(\bw_{\mathcal{T}})-F(\bw_1)\le G\mathcal{T} \eta=O(\log(T))
\end{align*}
so that $F(\bw_{\mathcal{T}}) - F(\bw_{T+1}) \le O(F(\bw_1)-F(\bw_{T+1}) + \log(T))$. Now consider two cases, either 
\begin{align*}
    \min_{t\ge \mathcal{T}} \|\bm_t\|_\star\le 2\left(\frac{6Z }{T^{\frac{\power-1}{3\power-2}}} + \frac{Ls}{2T^{\frac{2\power-1}{3\power-2}}}\right)
\end{align*}
or not. In the first case, by our bound on $\|\hat \epsilon_t\|_\star$, the desired bound on $\|\nabla F(\bw_t)\|_\star$ follows. In the latter case, we have
\begin{align*}
     F(\bw_{T+1})&\le F(\bw_{\mathcal{T}})-\frac{\eta}{2}\sum_{t=\mathcal{T}}^T \|\bm_t\|_\star
\end{align*}
So that 
\begin{align*}
    \min_{t\ge \mathcal{T}} \|\bm_t\|_\star \le O\left(\frac{F(\bw_1)-F(\bw_{T+1}) + \log(T)}{T\eta}\right) = \tilde O\left(\frac{1}{T^{\frac{2\power-1}{3\power-2}}}\right)
\end{align*}
And so again the result follows from our bound on $\|\hat\epsilon_t\|_\star$. 
\end{proof}

In addition to providing a method for identifying critical points, Theorem \ref{thm:nsgdclipburnin} also provides an intuitively desirable guarantee about the \emph{last-iterate} of the algorithm $\bw_T$. Specifically, one can easily check that the following Corollary:
\begin{Corollary}\label{thm:lastiterate}
Under the notation and assumptions of Theorem \ref{thm:nsgdclipburnin}, with probability at least $1-\delta$, there exists some $\hat w$ such that both of the following inequalities hold:
\begin{align*}
    \|\nabla F(\hat w)\|_\star&\le O\left(\frac{\log(T/\delta)}{T^{\frac{\power-1}{3\power-2}}}\right)\\
    F(\bw_T) &\le F(\hat w)
\end{align*}
where $\bw_T$ is the last iterate of Algorithm \ref{alg:nsgdclip}
\end{Corollary}
\begin{proof}
Set $\hat t$ to be the last index such that $\bw_{\hat t}$ does not satisfy (\ref{eqn:mcondition}). Set $\hat w = \bw_{\hat t}$. Then Theorem \ref{thm:nsgdclipburnin} implies the result.
\end{proof}

In most applications we are not actually interested in finding critical points, but instead wish to actually minimize the objective $F$. This observation tells us that our objective value is at least never worse than it would have been if had indeed searched specifically for a critical point. Moreover, by providing a guarantee about $\bw_T$, we have a closer match to practical use-cases: many of the theoretical analyses of non-convex stochastic gradient methods \citep{ghadimi2013stochastic, ward2019adagrad,li2019convergence} only show that a \emph{randomly selected iterate} has a small gradient norm. In contrast, Corollary \ref{thm:lastiterate} provides a non-asymptotic guarantee for the final iterate, which is the iterate most likely to be deployed after training in practice.

\section{Second-Order Smooth Losses}\label{sec:igt}
In this section, we provide our extension to losses that are \emph{second order smooth}. The overall method is very similar: we employ normalized SGD with momentum. However, in order to take advantage of the second-order smoothness, we will need to use a more advanced form of momentum. Specifically, we use implicit gradient transport \citep{arnold2019reducing}, as implemented by \citep{cutkosky2020momentum} in their NIGT algorithm. This algorithm replaces the standard momentum update with an extrapolation procedure:
\begin{align*}
    \bm_t = \beta \bm_{t-1} +  (1-\beta) \nabla f\left(\bw_t + \frac{\beta(\bw_t - \bw_{t-1})}{1-\beta}, z_t\right)
\end{align*}
By evaluating the gradient at the ``shifted'' point $\bw_t + \frac{\beta(\bw_t - \bw_{t-1})}{1-\beta}$ and performing a second-order Taylor expansion enabled by the second-order smoothness assumption, it is possible to show that $\bm_t$ is a less-biased estimate of $\nabla F(\bw_t)$ than it would be using the standard momentum update. We augment this procedure with gradient clipping in Algorithm \ref{alg:nsgdclipigt} below, and provide its analysis in Theorems \ref{thm:nsgdclipigt} and \ref{thm:nsgdclipburninigt}, which are directly analogous to Theorems \ref{thm:nsgdclip} and \ref{thm:nsgdclipburnin}.
\begin{algorithm}
   \caption{NIGT with Clipping}
   \label{alg:nsgdclipigt}
   \begin{algorithmic}
      \STATE{\bfseries Input: } Initial Point $\bw_1$, learning rate $\eta$, momentum parameter $\beta$, clipping parameter $\tau$, time horizon $T$:
      \STATE Set $\bm_0 = 0$.
      \FOR{$t=1\dots T$}
      \STATE Sample $z_t\sim P_z$.
      \STATE $\bx_t = \bw_t + \frac{\beta(\bw_t - \bw_{t-1})}{1-\beta}$
      \STATE Set $\bgclip_t = \frac{\nabla f(\bx_t, z_t)}{\|\nabla f(\bx_t, z_t)\|_\star}\min(\tau, \|\nabla f(\bx_1,z_t)\|_\star)$.
      \STATE 
      \STATE Set $\bm_t = \beta \bm_{t-1} +  (1-\beta) \bgclip_t$.
      \STATE Set $\bw_{t+1} = \bw_t  - \eta d(\bm_t)$.
      \ENDFOR
   \end{algorithmic}
\end{algorithm}

\begin{restatable}{Theorem}{thmnsgdclipigt}\label{thm:nsgdclipigt}
Suppose $\E_z[\|\nabla f(\bw,z)\|^\power]\le G^\power$ for all $\bw$ for some $G$. Suppose $F$ is $L$-smooth and $\rho$-second-order smooth. Set $\beta=1-\alpha$, $\alpha = \frac{b}{T^{\frac{2\power}{5\power-3}}}$ and $\eta = \frac{a}{T^{\frac{3\power -1}{5\power-3}}}$ for arbitrary constants $b$ and $c$ satisfying $\alpha \le 1$, and set $\tau=\frac{G}{\alpha^{1/\power}}$. Then with probability at least $1-\delta$:
\begin{align*}
    \frac{1}{T}\sum_{t=1}^T \|\nabla F(\bw_t)\|_\star&\le O\left(\frac{\log(T/\delta)}{ T^{\frac{2\power-2}{5\power-3}}}\right)
\end{align*}
where the big-Oh hides constant that depend on $L$, $\rho$, $G$, $b$, $s$, $C$, and $F(\bw_1)-F(\bw_{T+1})$, but not $T$ or $\delta$.
\end{restatable}

We also have a direct analog of Theorem \ref{thm:nsgdclipburnin}
\begin{restatable}{Theorem}{thmnsgdclipburninigt}\label{thm:nsgdclipburninigt}
Under the assumptions of Theorem \ref{thm:nsgdclipigt}, define the constants:
\begin{align*}
K&=10\max(1,\log(3T/\delta)) + 4\sqrt{C \max(1,\log(3T/\delta))} + 1\\
    Z&= \frac{\rho s^2}{b^2} + GK b^{\frac{\power-1}{\power}}
\end{align*}
Then, with probability at least $1-\delta$, if
\begin{align*}
t\ge \mathcal{T}=\frac{T^\frac{2\power}{5\power-3}}{b}\left(\frac{2\power-2}{5\power-3}\log(T) +\log(G)- \log(Z)\right)
\end{align*}
Then we have
\begin{align*}
    \|\bm_t-\nabla F(\bx_t)\|&\le\frac{2Z}{T^{\frac{2\power-2}{5\power-3}}}
\end{align*}
And so long as
\begin{align}
\|\bm_t\|_\star\ge 2\left(\frac{6Z }{T^{\frac{2\power-2}{5\power-3}}} + \frac{Ls}{2T^{\frac{3\power-1}{5\power-3}}}\right)\label{eqn:mconditionigt}
\end{align}
we have $F(\bw_{t+1})< F(\bw_t)-\frac{\eta}{2} \|\bm_t\|$. Moreover, if (\ref{eqn:mconditionigt}) is ever not satisfied, we must have 
\begin{align*}
    \|\nabla F(\bw_t)\|_\star\le \frac{14 Z}{T^{\frac{2\power-2}{5\power-3}}} + \frac{Ls}{T^{\frac{3\power-1}{5\power-3}}}
\end{align*}
Finally, if $t$ is the iteration with smallest value of $\|\bm_t\|$ such that $t\ge \mathcal{T}$, we have with probability at least $1-\delta$:
\begin{align*}
    \|\nabla F(\bw_t)\|_\star\le O\left(\frac{\log(T/\delta)}{T^{\frac{2\power-2}{5\power-3}}}\right)
\end{align*}
\end{restatable}

Finally, we note that we have a statement about the last iterate of Algorithm \ref{alg:nsgdclipigt} that is directly analogous to Corollary \ref{thm:lastiterate}:
\begin{Corollary}\label{thm:lastiterateigt}
Under the notation and assumptions of Theorem \ref{thm:nsgdclipburninigt}, with probability at least $1-\delta$, there exists some $\hat w$ such that both of the following inequalities hold:
\begin{align*}
    \|\nabla F(\hat w)\|_\star&\le O\left(\frac{\log(T/\delta)}{T^{\frac{2\power-2}{5\power-2}}}\right)\\
    F(\bw_T) &\le F(\hat w)
\end{align*}
where $\bw_T$ is the last iterate of Algorithm \ref{alg:nsgdclipigt}
\end{Corollary}

\begin{figure*}[h!]
\centering
          \begin{subfigure}
     \centering
     \includegraphics[width=0.45\textwidth]{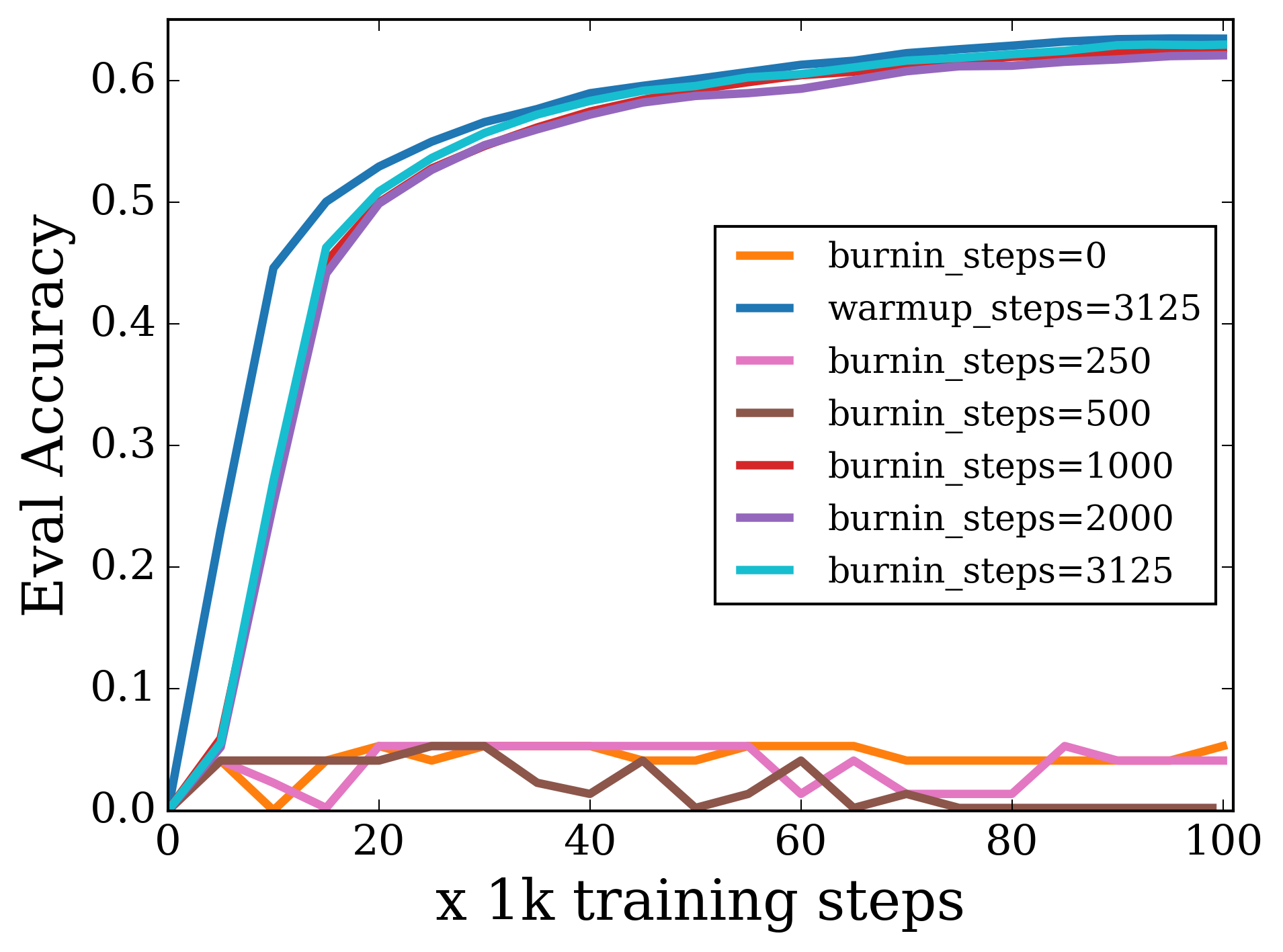}
     \label{fig:bert_lrage_alg1}
     \qquad
  \end{subfigure}
      \begin{subfigure}
     \centering
     \includegraphics[width=0.45\textwidth]{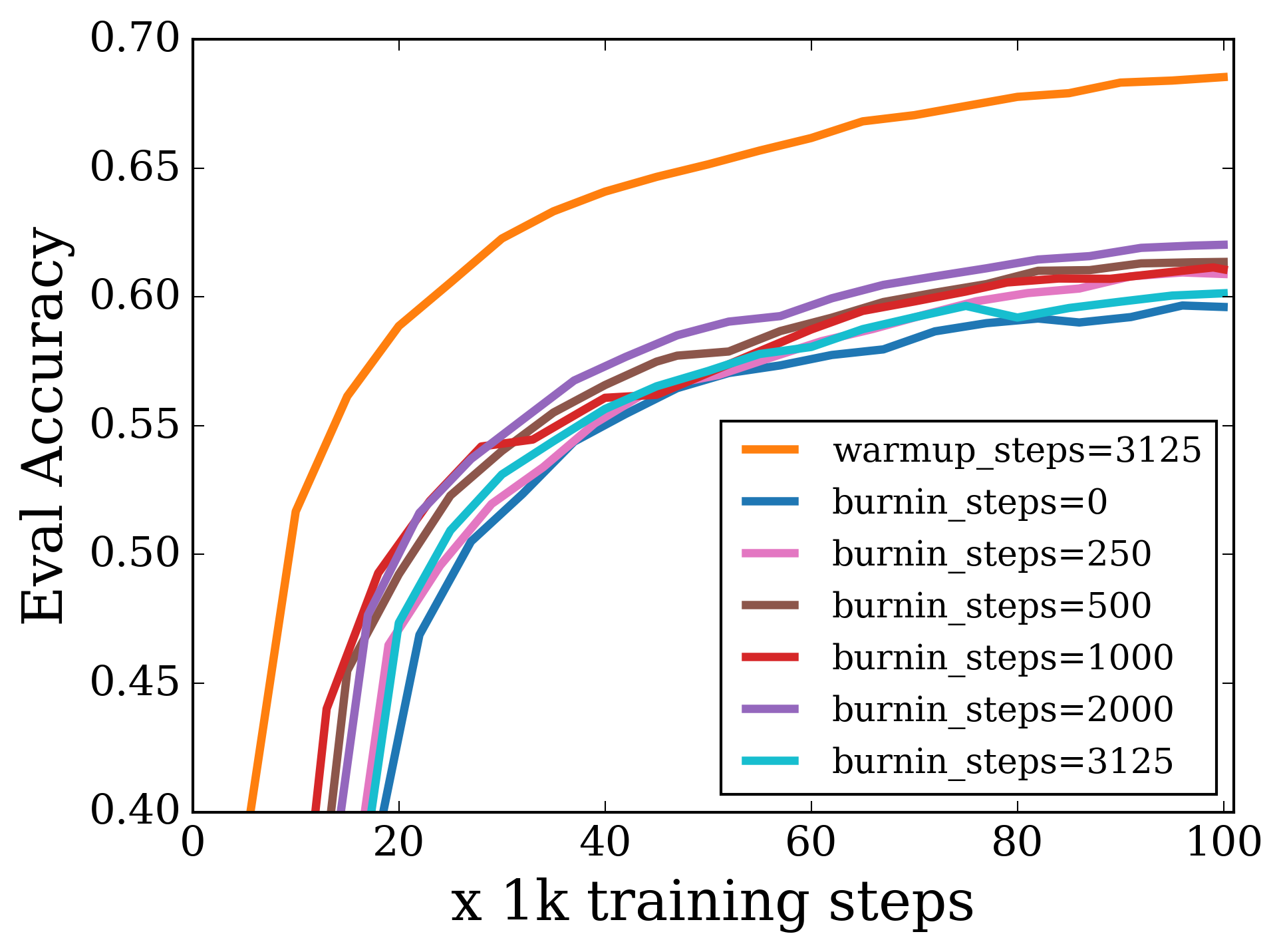}
     \label{fig:bert_lrage}
  \end{subfigure}
\caption{Comparison between warm-up and burn-in strategies on Masked language modeling BERT-Large pre-training task. Plot (a) shows the eval accuracy when trained with Algorithm \ref{alg:nsgdclip} and plot (b) does the same when trained with LAMB.}
  \label{fig:experiments}
\end{figure*}
\section{Experiments}\label{sec:experiments}
Our theoretical analysis suggests that there is a short ``burn-in'' phase in our algorithms during which the momentum estimate $\bm_t$ may be far from the true gradient $\nabla F(\bw_t)$, and after this period, $\bm_t$ will always be close to $\nabla F(\bw_t)$. As a result, during the burn-in phase it is possible for the optimizer to make slow or even negative progress, while afterwards we should expect more steady improvement. This suggests that one might set the learning rate $\eta=0$ throughout the burn-in phase and only begin taking steps after $\bm_t$ has accumulated enough gradients to be come a good estimate. This procedure is reminiscent of the common practice of learning rate ``warm-up'', in which the learning rate starts at 0 and is linearly increased to some desired value during the initial phase of training, and then later decreased. Thus, we sought to answer how much of the success of the warm-up schedule can be attributed to burn-in.

To answer this question, we conducted an experimental study on the BERT pretraining task. We chose to experiment on BERT since it has been shown to have heavy tails empirically \citep{zhang2020adaptive} and the common practice is already to clip the gradients. Figure \ref{fig:bert_lrage_alg1} plots the masked language model accuracy of BERT-Large model when trained with Algorithm \ref{alg:nsgdclip}. We also include experimental results using the LAMB optimizer instead of Algorithm \ref{alg:nsgdclip}. LAMB optimizer has been shown to obtain state of the art performance on BERT task \citep{You2020LargeLAMB}. Moreover, both our analysis and the LAMB optimizer relies on normalized updates. Note that we also include experimental results using the smaller BERT-Base model in the appendix. All our experiments were conducted using the Tensorflow framework on TPUv3 architecture. 

When using Algorithm \ref{alg:nsgdclip}, we employ base learning rate $\eta_{0}$ of 0.3 and batch size of 512. We came up with this learning rate by performing grid search $\eta_{0} \in [1.0, 0.5, 0.1, 0.05, 0.01, 0.005, 0.001, 0.0005, 0.0001]$ and choosing the one which attains the best eval accuracy. To obtain the optimal base learning rate of 0.3, we again ran a grid search $\eta_{0} \in [0.1, 0.2, 0.3, 0.4, 0.5]$. Our warmup baseline employs the standard practice of linear warm-up for 3125 steps and polynomical decay of $\eta_{t} = \eta_{0}*(1 - \frac{t}{T})$ for the rest of the steps. For a fair comparison with the baseline, we start the polynomial decay after 3125 steps regardless of how many steps were used for burn-in. We set the learning rate to the constant base learning rate value of 0.3 for BERT-Large for steps in between burn-in and polynomial decay. For the models trained using burn-in, instead of linear warm-up, we set the base learning rate $\eta_{0}$ to zero for the amount of steps we run the burn-in for. As shown in Fig \ref{fig:experiments}, we perform a grid search across $[0, 250, 500, 1000, 2000, 3125]$ steps for the initial burn-in period and compare their results. Notice that, without warm-up or burn-in (i.e. 0 burn-in steps), we see very poor performance. Further, this poor performance persists for smaller values of burn-in, but with large enough burn-in values we suddenly see very good results. Once this critical level of burn-in is achieved, we continue to see small gains as the amount of burn-in is increased. This suggests that without burn-in or warm-up, the first iterations are taking extremely poor steps in the non-convex landscape  that may destroy some favorable properties of the initialization point. With enough burn-in, each step is well-aligned with the true gradient and so we may avoid this bad behavior. Further, we see that employing burn-in recovers almost all of the advantage gained from employing warm-up (final accuracy of 62.95\% with burn-in vs. 63.45\% with warm-up).  These results lend empirical support to our theoretical intuition that the initial steps of the optimization may be unhelpful, or even hurtful. 


For the LAMB optimizer, we reproduce all the hyperparameters from \citet{You2020LargeLAMB}. Most importantly, we train all our models using the base learning rate $\eta_{0}$ of 6.25e-4 and batch size of 512. As shown in Fig \ref{fig:experiments}, again the warmup baseline employs the standard practice of linear warm-up for 3125 steps \citep{You2020LargeLAMB} and polynomical decay of $\eta_{t} = \eta_{0}*(1 - \frac{t}{T})$ for the rest of the steps. Again, for a fair comparison with the baseline, we start the polynomial decay after 3125 steps regardless of how many steps were used for burn-in.  We set the learning rate to the constant base learning rate value of 6.25e-4 for steps in between burn-in and polynomial decay. Note that when using the LAMB optimizer, warm-up baseline does considerably better than models where burn-in was employed. However, notice that the gap between employing warm-up and using \emph{neither} warm-up nor burn-in is considerably smaller than with  Algorithm~\ref{alg:nsgdclip}, for which not using either technique results in extremely poor accuracy. This suggests that the additional heuristics employed by the LAMB optimizer over the simple normalized SGD approach of Algorithm~\ref{alg:nsgdclip} are providing some non-trivial robustness that remains to be understood. Further, there is still a noticeable gap between the linear learning-rate increase in warm-up and the coarser burn-in policy. It remains to be seen exactly what mechanism this smoother policy employs to aid optimization.

\section{Conclusion}\label{sec:conclusion}

We have provided algorithms based on normalized SGD with momentum and gradient clipping that provide high probability-rates for finding critical points of non-convex objectives with heavy-tailed gradients. We also show that as a by-product of our high-probability bounds, the algorithms provide a pleasing result for the \emph{last iterate} of the optimization procedure. Specifically, the last iterate has function value smaller than the function value of a critical point.

Finally, we identify a critical ``burn-in'' phase early in optimization in which the optimizer may be making poor decisions. Empirically, we show that keeping the learning rate zero for the first iterations indeed does help training compared to simply running the optimizer with no special treatment during this phase. It is an interesting question to precisely identify what other advantages are provided by the increasing schedule used by warm-up during this phase.

\paragraph*{Acknowledgments:} The authors thank the anonymous reviewers for many helpful suggestions. AC was a visiting researcher at Google when this work was completed.
\bibliography{all}

\begin{thebibliography}{8}
\providecommand{\natexlab}[1]{#1}
\providecommand{\url}[1]{\texttt{#1}}
\expandafter\ifx\csname urlstyle\endcsname\relax
  \providecommand{\doi}[1]{doi: #1}\else
  \providecommand{\doi}{doi: \begingroup \urlstyle{rm}\Url}\fi

\bibitem[Author(2021)]{anonymous}
Author, N.~N.
\newblock Suppressed for anonymity, 2021.

\bibitem[Duda et~al.(2000)Duda, Hart, and Stork]{DudaHart2nd}
Duda, R.~O., Hart, P.~E., and Stork, D.~G.
\newblock \emph{Pattern Classification}.
\newblock John Wiley and Sons, 2nd edition, 2000.

\bibitem[Kearns(1989)]{kearns89}
Kearns, M.~J.
\newblock \emph{Computational Complexity of Machine Learning}.
\newblock PhD thesis, Department of Computer Science, Harvard University, 1989.

\bibitem[Langley(2000)]{langley00}
Langley, P.
\newblock Crafting papers on machine learning.
\newblock In Langley, P. (ed.), \emph{Proceedings of the 17th International
  Conference on Machine Learning (ICML 2000)}, pp.\  1207--1216, Stanford, CA,
  2000. Morgan Kaufmann.

\bibitem[Michalski et~al.(1983)Michalski, Carbonell, and
  Mitchell]{MachineLearningI}
Michalski, R.~S., Carbonell, J.~G., and Mitchell, T.~M. (eds.).
\newblock \emph{Machine Learning: An Artificial Intelligence Approach, Vol. I}.
\newblock Tioga, Palo Alto, CA, 1983.

\bibitem[Mitchell(1980)]{mitchell80}
Mitchell, T.~M.
\newblock The need for biases in learning generalizations.
\newblock Technical report, Computer Science Department, Rutgers University,
  New Brunswick, MA, 1980.

\bibitem[Newell \& Rosenbloom(1981)Newell and Rosenbloom]{Newell81}
Newell, A. and Rosenbloom, P.~S.
\newblock Mechanisms of skill acquisition and the law of practice.
\newblock In Anderson, J.~R. (ed.), \emph{Cognitive Skills and Their
  Acquisition}, chapter~1, pp.\  1--51. Lawrence Erlbaum Associates, Inc.,
  Hillsdale, NJ, 1981.

\bibitem[Samuel(1959)]{Samuel59}
Samuel, A.~L.
\newblock Some studies in machine learning using the game of checkers.
\newblock \emph{IBM Journal of Research and Development}, 3\penalty0
  (3):\penalty0 211--229, 1959.

\end{thebibliography}


\begin{thebibliography}{43}
\providecommand{\natexlab}[1]{#1}
\providecommand{\url}[1]{\texttt{#1}}
\expandafter\ifx\csname urlstyle\endcsname\relax
  \providecommand{\doi}[1]{doi: #1}\else
  \providecommand{\doi}{doi: \begingroup \urlstyle{rm}\Url}\fi

\bibitem[Allen-Zhu(2018)]{allen2018natasha}
Allen-Zhu, Z.
\newblock Natasha 2: Faster non-convex optimization than sgd.
\newblock In \emph{Advances in neural information processing systems}, pp.\
  2675--2686, 2018.

\bibitem[Arjevani et~al.(2019)Arjevani, Carmon, Duchi, Foster, Srebro, and
  Woodworth]{arjevani2019lower}
Arjevani, Y., Carmon, Y., Duchi, J.~C., Foster, D.~J., Srebro, N., and
  Woodworth, B.
\newblock Lower bounds for non-convex stochastic optimization.
\newblock \emph{arXiv preprint arXiv:1912.02365}, 2019.

\bibitem[Arjevani et~al.(2020)Arjevani, Carmon, Duchi, Foster, Sekhari, and
  Sridharan]{arjevani2020second}
Arjevani, Y., Carmon, Y., Duchi, J.~C., Foster, D.~J., Sekhari, A., and
  Sridharan, K.
\newblock Second-order information in non-convex stochastic optimization: Power
  and limitations.
\newblock In \emph{Conference on Learning Theory}, pp.\  242--299, 2020.

\bibitem[Arnold et~al.(2019)Arnold, Manzagol, Babanezhad, Mitliagkas, and
  Roux]{arnold2019reducing}
Arnold, S.~M., Manzagol, P.-A., Babanezhad, R., Mitliagkas, I., and Roux, N.~L.
\newblock Reducing the variance in online optimization by transporting past
  gradients.
\newblock \emph{arXiv preprint arXiv:1906.03532}, 2019.

\bibitem[Brown et~al.(2020)Brown, Mann, Ryder, Subbiah, Kaplan, Dhariwal,
  Neelakantan, Shyam, Sastry, Askell, Agarwal, Herbert-Voss, Krueger, Henighan,
  Child, Ramesh, Ziegler, Wu, Winter, Hesse, Chen, Sigler, Litwin, Gray, Chess,
  Clark, Berner, McCandlish, Radford, Sutskever, and Amodei]{brown2020language}
Brown, T.~B., Mann, B., Ryder, N., Subbiah, M., Kaplan, J., Dhariwal, P.,
  Neelakantan, A., Shyam, P., Sastry, G., Askell, A., Agarwal, S.,
  Herbert-Voss, A., Krueger, G., Henighan, T., Child, R., Ramesh, A., Ziegler,
  D.~M., Wu, J., Winter, C., Hesse, C., Chen, M., Sigler, E., Litwin, M., Gray,
  S., Chess, B., Clark, J., Berner, C., McCandlish, S., Radford, A., Sutskever,
  I., and Amodei, D.
\newblock Language models are few-shot learners, 2020.

\bibitem[Bubeck et~al.(2013)Bubeck, Cesa-Bianchi, and
  Lugosi]{bubeck2013bandits}
Bubeck, S., Cesa-Bianchi, N., and Lugosi, G.
\newblock Bandits with heavy tail.
\newblock \emph{IEEE Transactions on Information Theory}, 59\penalty0
  (11):\penalty0 7711--7717, 2013.

\bibitem[Chen et~al.(2018)Chen, Firat, Bapna, Johnson, Macherey, Foster, Jones,
  Schuster, Shazeer, Parmar, and et~al.]{Chen_2018_nmt_warmup}
Chen, M.~X., Firat, O., Bapna, A., Johnson, M., Macherey, W., Foster, G.,
  Jones, L., Schuster, M., Shazeer, N., Parmar, N., and et~al.
\newblock The best of both worlds: Combining recent advances in neural machine
  translation.
\newblock \emph{Proceedings of the 56th Annual Meeting of the Association for
  Computational Linguistics (Volume 1: Long Papers)}, 2018.
\newblock \doi{10.18653/v1/p18-1008}.
\newblock URL \url{http://dx.doi.org/10.18653/v1/P18-1008}.

\bibitem[Cutkosky(2018)]{cutkosky2018algorithms}
Cutkosky, A.
\newblock \emph{Algorithms and Lower Bounds for Parameter-free Online
  Learning}.
\newblock PhD thesis, Stanford University, 2018.

\bibitem[Cutkosky \& Mehta(2020)Cutkosky and Mehta]{cutkosky2020momentum}
Cutkosky, A. and Mehta, H.
\newblock Momentum improves normalized sgd.
\newblock \emph{arXiv preprint arXiv:2002.03305}, 2020.

\bibitem[Devlin et~al.(2019)Devlin, Chang, Lee, and
  Toutanova]{devlin-etal-2019-bert}
Devlin, J., Chang, M.-W., Lee, K., and Toutanova, K.
\newblock {BERT}: Pre-training of deep bidirectional transformers for language
  understanding.
\newblock In \emph{Proceedings of the 2019 Conference of the North {A}merican
  Chapter of the Association for Computational Linguistics: Human Language
  Technologies, Volume 1 (Long and Short Papers)}, pp.\  4171--4186,
  Minneapolis, Minnesota, June 2019. Association for Computational Linguistics.
\newblock \doi{10.18653/v1/N19-1423}.
\newblock URL \url{https://www.aclweb.org/anthology/N19-1423}.

\bibitem[Duchi et~al.(2010)Duchi, Hazan, and Singer]{duchi10adagrad}
Duchi, J., Hazan, E., and Singer, Y.
\newblock Adaptive subgradient methods for online learning and stochastic
  optimization.
\newblock In \emph{Conference on Learning Theory (COLT)}, 2010.

\bibitem[Fang et~al.(2019)Fang, Lin, and Zhang]{fang2019sharp}
Fang, C., Lin, Z., and Zhang, T.
\newblock Sharp analysis for nonconvex sgd escaping from saddle points.
\newblock In \emph{Conference on Learning Theory}, pp.\  1192--1234, 2019.

\bibitem[Ghadimi \& Lan(2013)Ghadimi and Lan]{ghadimi2013stochastic}
Ghadimi, S. and Lan, G.
\newblock Stochastic first-and zeroth-order methods for nonconvex stochastic
  programming.
\newblock \emph{SIAM Journal on Optimization}, 23\penalty0 (4):\penalty0
  2341--2368, 2013.

\bibitem[Gorbunov et~al.(2020)Gorbunov, Danilova, and
  Gasnikov]{gorbunov2020stochastic}
Gorbunov, E., Danilova, M., and Gasnikov, A.
\newblock Stochastic optimization with heavy-tailed noise via accelerated
  gradient clipping.
\newblock In \emph{Neural Information Processing Systems}, 2020.

\bibitem[Goyal et~al.(2017)Goyal, Dollár, Girshick, Noordhuis, Wesolowski,
  Kyrola, Tulloch, Jia, and He]{goyal2017accurate}
Goyal, P., Dollár, P., Girshick, R., Noordhuis, P., Wesolowski, L., Kyrola,
  A., Tulloch, A., Jia, Y., and He, K.
\newblock Accurate, large minibatch sgd: Training imagenet in 1 hour, 2017.

\bibitem[He et~al.(2015)He, Zhang, Ren, and Sun]{He2015DeepRL}
He, K., Zhang, X., Ren, S., and Sun, J.
\newblock Deep residual learning for image recognition.
\newblock \emph{2016 IEEE Conference on Computer Vision and Pattern Recognition
  (CVPR)}, pp.\  770--778, 2015.

\bibitem[Howard et~al.(2018)Howard, Ramdas, McAuliffe, and
  Sekhon]{howard2018time}
Howard, S.~R., Ramdas, A., McAuliffe, J., and Sekhon, J.
\newblock Time-uniform, nonparametric, nonasymptotic confidence sequences.
\newblock \emph{arXiv preprint arXiv:1810.08240}, 2018.

\bibitem[Huang et~al.(2020)Huang, Perez, Ba, and Volkovs]{pmlr-v119-huang20f}
Huang, X.~S., Perez, F., Ba, J., and Volkovs, M.
\newblock Improving transformer optimization through better initialization.
\newblock In III, H.~D. and Singh, A. (eds.), \emph{Proceedings of the 37th
  International Conference on Machine Learning}, volume 119 of
  \emph{Proceedings of Machine Learning Research}, pp.\  4475--4483. PMLR,
  13--18 Jul 2020.
\newblock URL \url{http://proceedings.mlr.press/v119/huang20f.html}.

\bibitem[Kingma \& Ba(2014)Kingma and Ba]{kingma2014adam}
Kingma, D. and Ba, J.
\newblock Adam: A method for stochastic optimization.
\newblock \emph{arXiv preprint arXiv:1412.6980}, 2014.

\bibitem[Li \& Orabona(2019)Li and Orabona]{li2019convergence}
Li, X. and Orabona, F.
\newblock On the convergence of stochastic gradient descent with adaptive
  stepsizes.
\newblock In \emph{The 22nd International Conference on Artificial Intelligence
  and Statistics}, pp.\  983--992. PMLR, 2019.

\bibitem[Li \& Orabona(2020)Li and Orabona]{li2020high}
Li, X. and Orabona, F.
\newblock A high probability analysis of adaptive sgd with momentum.
\newblock \emph{arXiv preprint arXiv:2007.14294}, 2020.

\bibitem[Liu et~al.(2020)Liu, Jiang, He, Chen, Liu, Gao, and Han]{Liu2020On}
Liu, L., Jiang, H., He, P., Chen, W., Liu, X., Gao, J., and Han, J.
\newblock On the variance of the adaptive learning rate and beyond.
\newblock In \emph{International Conference on Learning Representations}, 2020.
\newblock URL \url{https://openreview.net/forum?id=rkgz2aEKDr}.

\bibitem[Lugosi \& Mendelson(2019)Lugosi and Mendelson]{lugosi2019mean}
Lugosi, G. and Mendelson, S.
\newblock Mean estimation and regression under heavy-tailed distributions: A
  survey.
\newblock \emph{Foundations of Computational Mathematics}, 19\penalty0
  (5):\penalty0 1145--1190, 2019.

\bibitem[Minsker(2011)]{minsker2011some}
Minsker, S.
\newblock On some extensions of bernstein's inequality for self-adjoint
  operators.
\newblock \emph{arXiv preprint arXiv:1112.5448}, 2011.

\bibitem[Nguyen \& Salazar(2019)Nguyen and Salazar]{Nguyen2019TransformersWT}
Nguyen, T.~Q. and Salazar, J.
\newblock Transformers without tears: Improving the normalization of
  self-attention.
\newblock \emph{ArXiv}, abs/1910.05895, 2019.

\bibitem[Pascanu et~al.(2012)Pascanu, Mikolov, and
  Bengio]{pascanu2012understanding}
Pascanu, R., Mikolov, T., and Bengio, Y.
\newblock Understanding the exploding gradient problem.
\newblock \emph{CoRR, abs/1211.5063}, 2\penalty0 (417):\penalty0 1, 2012.

\bibitem[Raffel et~al.(2020)Raffel, Shazeer, Roberts, Lee, Narang, Matena,
  Zhou, Li, and Liu]{JMLR:v21:20-074:T5-paper}
Raffel, C., Shazeer, N., Roberts, A., Lee, K., Narang, S., Matena, M., Zhou,
  Y., Li, W., and Liu, P.~J.
\newblock Exploring the limits of transfer learning with a unified text-to-text
  transformer.
\newblock \emph{Journal of Machine Learning Research}, 21\penalty0
  (140):\penalty0 1--67, 2020.
\newblock URL \url{http://jmlr.org/papers/v21/20-074.html}.

\bibitem[Reddi et~al.(2018)Reddi, Kale, and Kumar]{reddi2018on}
Reddi, S.~J., Kale, S., and Kumar, S.
\newblock On the convergence of adam and beyond.
\newblock In \emph{International Conference on Learning Representations}, 2018.

\bibitem[Scaman \& Malherbe(2020)Scaman and Malherbe]{scaman2020robustness}
Scaman, K. and Malherbe, C.
\newblock Robustness analysis of non-convex stochastic gradient descent using
  biased expectations.
\newblock \emph{Advances in Neural Information Processing Systems}, 33, 2020.

\bibitem[{\c{S}}im{\c{s}}ekli et~al.(2019){\c{S}}im{\c{s}}ekli,
  G{\"u}rb{\"u}zbalaban, Nguyen, Richard, and Sagun]{simcsekli2019heavy}
{\c{S}}im{\c{s}}ekli, U., G{\"u}rb{\"u}zbalaban, M., Nguyen, T.~H., Richard,
  G., and Sagun, L.
\newblock On the heavy-tailed theory of stochastic gradient descent for deep
  neural networks.
\newblock \emph{arXiv preprint arXiv:1912.00018}, 2019.

\bibitem[Simsekli et~al.(2019)Simsekli, Sagun, and
  Gurbuzbalaban]{simsekli2019tail}
Simsekli, U., Sagun, L., and Gurbuzbalaban, M.
\newblock A tail-index analysis of stochastic gradient noise in deep neural
  networks.
\newblock In \emph{International Conference on Machine Learning}, pp.\
  5827--5837. PMLR, 2019.

\bibitem[Tripuraneni et~al.(2018)Tripuraneni, Stern, Jin, Regier, and
  Jordan]{tripuraneni2018stochastic}
Tripuraneni, N., Stern, M., Jin, C., Regier, J., and Jordan, M.~I.
\newblock Stochastic cubic regularization for fast nonconvex optimization.
\newblock In \emph{Advances in neural information processing systems}, pp.\
  2899--2908, 2018.

\bibitem[Tropp(2011)]{tropp2011freedman}
Tropp, J.
\newblock Freedman's inequality for matrix martingales.
\newblock \emph{Electronic Communications in Probability}, 16:\penalty0
  262--270, 2011.

\bibitem[Tropp(2015)]{tropp2015introduction}
Tropp, J.~A.
\newblock An introduction to matrix concentration inequalities.
\newblock \emph{arXiv preprint arXiv:1501.01571}, 2015.

\bibitem[Vaswani et~al.(2017)Vaswani, Shazeer, Parmar, Uszkoreit, Jones, Gomez,
  Kaiser, and Polosukhin]{vaswani2017attention}
Vaswani, A., Shazeer, N., Parmar, N., Uszkoreit, J., Jones, L., Gomez, A.~N.,
  Kaiser, {\L}., and Polosukhin, I.
\newblock Attention is all you need.
\newblock \emph{Advances in Neural Information Processing Systems},
  30:\penalty0 5998--6008, 2017.

\bibitem[Wang et~al.(2021)Wang, G{\"u}rb{\"u}zbalaban, Zhu,
  {\c{S}}im{\c{s}}ekli, and Erdogdu]{wang2021convergence}
Wang, H., G{\"u}rb{\"u}zbalaban, M., Zhu, L., {\c{S}}im{\c{s}}ekli, U., and
  Erdogdu, M.~A.
\newblock Convergence rates of stochastic gradient descent under infinite noise
  variance.
\newblock \emph{arXiv preprint arXiv:2102.10346}, 2021.

\bibitem[Wang et~al.(2019)Wang, Li, Xiao, Zhu, Li, Wong, and Chao]{Wang_2019}
Wang, Q., Li, B., Xiao, T., Zhu, J., Li, C., Wong, D.~F., and Chao, L.~S.
\newblock Learning deep transformer models for machine translation.
\newblock \emph{Proceedings of the 57th Annual Meeting of the Association for
  Computational Linguistics}, 2019.
\newblock \doi{10.18653/v1/p19-1176}.
\newblock URL \url{http://dx.doi.org/10.18653/v1/P19-1176}.

\bibitem[Ward et~al.(2019)Ward, Wu, and Bottou]{ward2019adagrad}
Ward, R., Wu, X., and Bottou, L.
\newblock Adagrad stepsizes: Sharp convergence over nonconvex landscapes.
\newblock In \emph{International Conference on Machine Learning}, pp.\
  6677--6686. PMLR, 2019.

\bibitem[You et~al.(2017)You, Gitman, and Ginsburg]{you2017large}
You, Y., Gitman, I., and Ginsburg, B.
\newblock Large batch training of convolutional networks.
\newblock \emph{arXiv preprint arXiv:1708.03888}, 2017.

\bibitem[You et~al.(2020)You, Li, Reddi, Hseu, Kumar, Bhojanapalli, Song,
  Demmel, Keutzer, and Hsieh]{You2020LargeLAMB}
You, Y., Li, J., Reddi, S., Hseu, J., Kumar, S., Bhojanapalli, S., Song, X.,
  Demmel, J., Keutzer, K., and Hsieh, C.-J.
\newblock Large batch optimization for deep learning: Training bert in 76
  minutes.
\newblock In \emph{International Conference on Learning Representations}, 2020.
\newblock URL \url{https://openreview.net/forum?id=Syx4wnEtvH}.

\bibitem[Zhang et~al.(2020{\natexlab{a}})Zhang, Jin, Fang, and
  Wang]{zhang2020improved}
Zhang, B., Jin, J., Fang, C., and Wang, L.
\newblock Improved analysis of clipping algorithms for non-convex optimization.
\newblock \emph{arXiv preprint arXiv:2010.02519}, 2020{\natexlab{a}}.

\bibitem[Zhang et~al.(2019)Zhang, He, Sra, and Jadbabaie]{zhang2019gradient}
Zhang, J., He, T., Sra, S., and Jadbabaie, A.
\newblock Why gradient clipping accelerates training: A theoretical
  justification for adaptivity.
\newblock In \emph{International Conference on Learning Representations}, 2019.

\bibitem[Zhang et~al.(2020{\natexlab{b}})Zhang, Karimireddy, Veit, Kim, Reddi,
  Kumar, and Sra]{zhang2020adaptive}
Zhang, J., Karimireddy, S.~P., Veit, A., Kim, S., Reddi, S., Kumar, S., and
  Sra, S.
\newblock Why are adaptive methods good for attention models?
\newblock \emph{Advances in Neural Information Processing Systems}, 33,
  2020{\natexlab{b}}.

\end{thebibliography}
\bibliographystyle{icml2021}

\clearpage
\appendix

\section{Proofs for Section \ref{sec:igt}}

In this section we provide the missing proofs of Theorems \ref{thm:nsgdclipigt} and \ref{thm:nsgdclipburninigt}.

\thmnsgdclipigt*

\begin{proof}
The start of our proof is very similar to that of Theorem 3 in \citep{cutkosky2020momentum}.
Similar to the proof of Theorem \ref{thm:nsgdclip}, define $\hat\epsilon_t = \bm_t - \nabla F(\bw_t)$ and $\epsilon_t=\bgclip_t - \nabla F(\bx_t)$. Also (and as a departure from Theorem \ref{thm:nsgdclip}), define $S(a,b) = \nabla F(a)-\nabla F(b) - \nabla^2F(b)(a-b)$. Then we have a recursion for all $t\ge 1$:
\begin{align*}
    \bm_{t+1} &= (1-\alpha)(\nabla F(\bw_t) + \hat\epsilon_t) + \alpha \bgclip_{t+1}\\
    &=  (1-\alpha)(\nabla F(\bw_{t+1})+\nabla^2F(\bw_{t+1})(\bw_t-\bw_{t+1}) )+(1-\alpha)(S(\bw_t,\bw_{t+1})+ \hat\epsilon_t)\\
    &\quad+\alpha(\nabla F(\bw_{t+1}) + \nabla^2F(\bw_{t+1})(\bx_{t+1}-\bw_{t+1})+\alpha(S(\bx_{t+1},\bw_{t+1}) +\epsilon_{t+1})
    \intertext{Now, observe that by definition of $\bx_{t+1}$, the hessian-dependent terms cancel:}
    &=\nabla F(\bw_{t+1}) + (1-\alpha)(S(\bw_t,\bw_{t+1}) + \hat\epsilon_t)+\alpha( S(\bx_{t+1},\bw_{t+1}) + \epsilon_{t+1})\\
    \hat \epsilon_{t+1} &= (1-\alpha)(S(\bw_t,\bw_{t+1}) + \hat\epsilon_t)+\alpha( S(\bx_{t+1},\bw_{t+1}) + \epsilon_{t+1})
    \end{align*}
    Further, as in the proof of Theorem \ref{thm:nsgdclip}, we set $\bw_0=\bw_1$ and notice $\hat \epsilon_0 = -\nabla F(\bw_1)$ so that the above recursion holds for $t=0$ as well. Then we again unwind the recursion:
    \begin{align*}
    \hat \epsilon_{t+1} &= (1-\alpha)^{t+1}\hat \epsilon_0 + \alpha \sum_{t'=0}^t (1-\alpha)^{t'} \epsilon_{t+1-t'}+(1-\alpha)\sum_{t'=0}^t(1-\alpha)^{t'}S(\bw_{t-t'},\bw_{t-t'+1})\\
    &\quad+\alpha\sum_{t'=0}^t(1-\alpha)^{t'}S(\bx_{t-t'+1},\bw_{t-t'+1})\\
    \|\hat\epsilon_{t+1}\|_\star&\le (1-\alpha)^{t+1} \|\hat\epsilon_0\|_\star + \alpha \left\|\sum_{t'=0}^t (1-\alpha)^{t'} \epsilon_{t+1-t'}\right\|_\star+(1-\alpha)\left\|\sum_{t'=0}^t(1-\alpha)^{t'}S(\bw_{t-t'},\bw_{t-t'+1})\right\|_\star\\
    &\quad+\alpha\left\|\sum_{t'=0}^t(1-\alpha)^{t'}S(\bx_{t-t'+1},\bw_{t-t'+1})\right\|_\star
    \intertext{Now, by second-order smoothness, we have $\|S(a,b)\|_\star\le \frac{\rho}{2}\|a-b\|^2$ (see Proposition \ref{prop:secondsmoothgrad}) and further $\frac{(1-\alpha)^2}{\alpha} + 1-\alpha \le 2\frac{1-\alpha}{\alpha}$, so that combining the last two sums yields:}
    &\le  (1-\alpha)^{t+1} \|\hat\epsilon_0\|_\star + \alpha \left\|\sum_{t'=0}^t (1-\alpha)^{t'} \epsilon_{t+1-t'}\right\|_\star+\frac{(1-\alpha)\rho\eta^2}{\alpha }\sum_{t'=0}^t(1-\alpha)^{t'}\\
    &\le  (1-\alpha)^{t+1} \|\hat\epsilon_0\|_\star + \alpha \left\|\sum_{t'=0}^t (1-\alpha)^{t'} \epsilon_{t+1-t'}\right\|_\star+\frac{(1-\alpha)\rho\eta^2}{\alpha^2 }
\end{align*}
Now, by exactly the same argument as in Theorem \ref{thm:nsgdclip}, defining $D_\delta=\max(1,\log(1/\delta))$ we have with probability at least $1-\delta/T$:
\begin{align}
    \|\hat \epsilon_{t+1}\|_\star&\le (1-\alpha)^{t+1} G + \frac{(1-\alpha)\rho\eta^2}{\alpha^2 }+ 10C\alpha\tau D_\delta+ 4\sqrt{ \alpha CG^\power\tau^{2-\power}D_\delta}+\frac{G^\power}{ \tau^{\power-1}}\nonumber
    \intertext{
Then, with $\tau=\frac{G}{\alpha^{1/\power}}$, this becomes:}
&\le (1-\alpha)^{t+1} G + \frac{(1-\alpha)\rho\eta^2}{\alpha^2}+G\alpha^{1-1/\power}\left[10CD_\delta + 4\sqrt{C D_\delta} + 1\right]\nonumber\\
    &=(1-\alpha)^{t+1}G+ \frac{(1-\alpha)\rho\eta^2}{\alpha^2} + G\alpha^{1-1/\power}K\label{eqn:hatepsboundigt}
\end{align}
where we again define the constant $K=10D_\delta + 4\sqrt{C D_\delta} + 1$.

Now, we have from Lemma \ref{lem:onestep}:
\begin{align*}
    \sum_{t=1}^T \|\nabla F(\bw_t)\|_\star&\le \frac{F(\bw_1)-F(\bw_T)}{\eta } + \frac{LT\eta}{2}+2\sum_{t=1}^T \|\hat\epsilon_t\|_\star
    \intertext{so with probability at least $1-\delta$:}
    &\le \frac{F(\bw_1)-F(\bw_T)}{\eta } + \frac{LT\eta}{2}+2\sum_{t=1}^T\left[(1-\alpha)^{t+1}G+ \frac{(1-\alpha)\rho\eta^2}{\alpha^2}+ G\alpha^{1-1/\power}K\right]\\
    &\le \frac{F(\bw_1)-F(\bw_T)}{\eta } + \frac{LT\eta}{2}+\frac{2G(1-\alpha)}{\alpha} +\frac{2(1-\alpha)T\rho\eta^2}{\alpha^2} + 2GT\alpha^{1-1/\power}K
\end{align*}
Now set $\alpha = \frac{b}{T^{\frac{2\power}{5\power-3}}}$ and $\eta = \frac{s}{T^{\frac{3\power -1}{5\power-3}}}$. This yields with probability at least $1-\delta$:
\begin{align*}
    \frac{1}{T}\sum_{t=1}^T \|\nabla F(\bw_t)\|_\star&\le O\left( T^{\frac{2-2\power}{5\power-3}}\log(T/\delta)\right)
\end{align*}

\end{proof}

\thmnsgdclipburninigt*
\begin{proof}
Use the settings $\eta=\frac{s}{T^{\frac{3\power-1}{5\power-3}}}$ and $\alpha = \frac{b}{T^{\frac{2\power}{5\power-3}}}$ and notice that from equation (\ref{eqn:hatepsboundigt}) that with probability $1-\delta$, for all $t$:
\begin{align*}
    \|\hat \epsilon_{t+1}\|_\star
    &\le (1-\alpha)^tG+ \frac{(1-\alpha)\rho\eta^2}{\alpha^2} + G\alpha^{1-1/\power}K\\
    &=(1-\alpha)^t G + \frac{\rho s^2}{b^2T^{\frac{2\power-2}{5\power-3}}} + \frac{GK b^{\frac{\power-1}{\power}}}{T^{\frac{2\power-2}{5\power-3}}}\\
    &=(1-\alpha)^t G + \frac{Z}{T^{\frac{2\power-2}{5\power-3}}}
\end{align*}

Again, we have the identity:
\begin{align*}
    \log(1-\alpha)&\le -\alpha\\
    \log\left(G(1-\alpha)^t)\right) &\le -t\alpha +\log(G)
\end{align*}
Therefore, if $t$ satisfies:
\begin{align*}
    t\ge \frac{T^\frac{2\power}{5\power-3}}{b}\left(\frac{2\power-2}{5\power-3}\log(T) +\log(G) - \log\left(Z\right)\right)
\end{align*}
we have with probability at least $1-\delta$:
\begin{align*}
\|\hat \epsilon_{t+1}\|_\star&\le \frac{2Z}{T^{\frac{2\power-2}{5\power-3}}}
\end{align*}
Next, again from Lemma \ref{lem:onestep}:
\begin{align*}
    F(\bw_{t+1})&\le F(\bw_t) - \eta\|\nabla F(\bw_t)\|_\star + 2\eta\|\hat\epsilon_t\|_\star + \frac{L\eta^2}{2}\\
    &\le F(\bw_t) - \eta\|\bm_t\| + 3\eta\|\hat\epsilon_t\|_\star + \frac{L\eta^2}{2}\\
    &\le F(\bw_t) - \eta\|\bm_t\| +\eta\left(\frac{6Z }{T^{\frac{2\power-2}{5\power-3}}} + \frac{Ls}{2T^{\frac{3\power-1}{5\power-3}}}\right)
\end{align*}

Now, analogously to Theorem \ref{thm:nsgdclipburnin}, we argue that if $\|\bm_t\|\ge 2\left(\frac{6Z }{T^{\frac{2\power-2}{5\power-3}}} + \frac{Ls}{2T^{\frac{3\power-1}{5\power-3}}}\right)$, then $F(\bw_{t+1})\le F(\bw_t) - \frac{\eta}{2}\|\bm_t\|$.

The last part of the Theorem follows from an identical case-work argument to that of Theorem \ref{thm:nsgdclipburnin}.
\end{proof}
\section{A Concentration Bound for Truncated Random Vectors}\label{sec:concentration}
A key tool in proving our bounds is the following result, which provides a dimension-free concentration bound analogous to the concentration bounds for scalar random variables that can be found in \cite{lugosi2019mean, bubeck2013bandits}. We present the simpler results for Hilbert space first, and then generalize to the Banach space case.

\begin{restatable}{Lemma}{thmthresholdsimple}\label{thm:thresholdsimple}
Suppose $X_1,\dots,X_T$ are random vectors in a Hilbert space adapted to a filtration $\mathcal{F}_1, \mathcal{F}_2,\dots$ with means $\E_{t-1}[X_t]=\mu_t$. Suppose there is some $\power\in(1,2]$ such that for all $t$ there is some $G_t<\infty$ such that $\E[\|X_t\|^\power]\le G_t^\power$. Let $b_1,\dots,b_T$ be fixed constants, with $B=\max_t b_t$ such that $B\le 1$. For some $\tau>0$, let $\hat X_t =  \frac{X_t}{\|X_t\|}\min(\tau, \|X_t\|)$. Then with probability at least $1-\delta$:
\begin{align*}
    \left\|\sum_{t=1}^T b_t(\hat X_t - \mu_t)\right\|\le 4B\tau\log(3/\delta) +\sum_{t=1}^T\frac{b_tG_t^\power}{\tau^{\power-1}}+ 2\sqrt{\sum_{t=1}^T b_t^2G_t^\power \tau^{2-\power}\max(1,\log(3/\delta))}
\end{align*}
\end{restatable}
The proof has two main steps: first, we bound the bias and the variance of the truncated random vectors in terms of the truncation threhsold. Then, we use a dimension-free Feedman-style bound to show that the average of these truncated vectors concentrates about the mean of the original un-truncated vectors.

\begin{proof}
First, we will bound $\|\E[\hat X_t] - \mu_t\|$, following analysis similar to that present in \cite{zhang2020adaptive}:
\begin{align*}
    \|\E[\hat X_t] - \mu_t\|&=\|\E[\hat X_t -X_t]\|\\
    &\le \E[\|\hat X_t  - X_t \|]\\
    &\le \E[\|X_t\|\one[\|X_t\|\ge \tau]]\\
    &\le \E[\|X_t\|^\power/\tau^{\power-1}]\\
    &\le \frac{G_t^\power}{\tau^{\power-1}}
\end{align*}

Next, we bound the variance $\E[\|\hat X_t - \E[\hat X_t]\|^2]$:
\begin{align*}
    \E[\|\hat X_t - \E[\hat X_t]\|^2]&\le \E[\|\hat X_t\|^2]\\
    &\le \E[\|X_t\|^\power \tau^{2-\power}]\\
    &\le G_t^\power\tau^{2-\power}
\end{align*}

Now, invoke a vector-valued Feedman-inequality (Lemma \ref{thm:feedmanbanach}), observing that $\|\hat X_t\|\le \tau$ always so that $\|b_t(\hat X_t - \E[\hat X_t])\|\le 2b_t\tau\le 2B\tau$ where $B=\max_t b_t$. Thus, with probability $1-\delta$:

\begin{align*}
    \left\|\sum_{t=1}^T b_t(\hat X_t - \E[\hat X_t])\right\|\le 6B\tau\max(1,\log(3/\delta)) + 3\sqrt{\sum_{t=1}^T b_t^2G_t^\power \tau^{2-\power}\max(1,\log(3/\delta))}
\end{align*}
Therefore by triangle inequality:
\begin{align*}
    \left\|\sum_{t=1}^T b_t(\hat X_t - \mu_t)\right\|\le  6B\tau\max(1,\log(3/\delta)) +\sum_{t=1}^T\frac{b_tG_t^\power}{\tau^{\power-1}}+ 3\sqrt{\sum_{t=1}^T b_t^2G_t^\power \tau^{2-\power}\max(1,\log(3/\delta))}
\end{align*}

\end{proof}

Now, we present the generalization to Banach spaces.

\begin{restatable}{Lemma}{thmthresholdsimplebanach}\label{thm:thresholdsimplebanach}
Suppose $X_1,\dots,X_T$ are random vectors in some Banach space $\B$ satisfying (\ref{eqn:banach}) with $p\le2$ and $C\ge 1$ adapted to a filtration $F_1,F_2,\dots$. Let $\E_{t-1}[X_t]=\mu_t$. Suppose there is some $\power\in(1,2]$ such that for all $t$ there is some $G_t<\infty$ such that $\E[\|X_t\|^\power]\le G_t^\power$. Let $b_1,\dots,b_T$ be fixed constants, with $B=\max_t b_t$ such that $B\le 1$. For some $\tau>0$, let $\hat X_t =  \frac{X_t}{\|X_t\|}\min(\tau, \|X_t\|)$. Then with probability at least $1-\delta$:
\begin{align*}
    \left\|\sum_{t=1}^T b_t(\hat X_t - \mu_t)\right\|\le  10CB\tau\max(1,\log(3/\delta)) + \sum_{t=1}^T\frac{b_tG_t^\power}{\tau^{\power-1}} +4\left(C\sum_{t=1}^T b_t^p G_t^\power \tau^{p-\power}\right)^{1/p}\sqrt{\max(1,\log(3/\delta))}
\end{align*}
\end{restatable}
The proof follows the same structure as the previous result for Hilbert spaces.

\begin{proof}
Note that the bound
\begin{align*}
    \|\E[\hat X_t] - \mu_t\|&\le \frac{G_t^\power}{\tau^{\power-1}}
\end{align*}
continues to hold as the arguments apply without modification in Banach spaces. For the variance, we have an analogous statement:

\begin{align*}
    \E[\|\hat X_t - \E[\hat X_t]\|^p]&\le \E[\|\hat X_t\|^p]\\
    &\le \E[\|X_t\|^\power \tau^{p-\power}]\\
    &\le G_t^\power\tau^{p-\power}
\end{align*}

Now, invoke a vector-valued Feedman-inequality (Lemma \ref{thm:feedmanbanach}), observing that $\|\hat X_t\|\le \tau$ always so that $\|b_t(\hat X_t - \E[\hat\mu_t])\|\le 2b_t\tau\le 2B\tau$ where $B=\max_t b_t$. Thus, with probability $1-\delta$:

\begin{align*}
    \left\|\sum_{t=1}^T b_t(\hat X_t - \E[\hat X_t])\right\|\le 10CB\tau\max(1,\log(3/\delta)) +4\left(C\sum_{t=1}^k b_t^p G_t^\power \tau^{p-\power}\right)^{1/p}\sqrt{\max(1,\log(3/\delta)}
\end{align*}
Therefore by triangle inequality:
\begin{align*}
    \left\|\sum_{t=1}^T b_t(\hat X_t - \mu_t)\right\|\le 10CB\tau\max(1,\log(3/\delta)) + \sum_{t=1}^T\frac{b_tG_t^\power}{\tau^{\power-1}} +4\left(C\sum_{t=1}^k b_t^p G_t^\power \tau^{p-\power}\right)^{1/p}\sqrt{\max(1,\log(3/\delta))}
\end{align*}

\end{proof}

\section{Dimension-Free Martingale concentration from 1-d concentration}

In this section, we show how to obtain a dimension-free Freedman-style martingale concentration bound in a Banach space via a reduction to the ordinary 1-dimensional Freedman inequality. The conversion is based on a construction outlined in \cite{cutkosky2018algorithms} for converting 1-dimensional online linear optimization algorithms into dimension-free algorithms. It may be of independent interest as simple alternative proof for such concentration results, as well as providing a possibly more user-friendly result in comparison to advanced bounds available in \cite{tropp2015introduction, minsker2011some, howard2018time}. However, we emphasis that the results in this section are essentially provided only for completeness: they do \emph{not} improve upon previously available bounds.

Suppose $B$ is a real Banach space whose norm $\|\cdot\|$ is Frechet differentiable and satisfies for some $p$ and $C$ for all $x,y$:
\begin{align}
    \|x+y\|^p\le \|x\|^p+\langle \nabla \|x\|^p,y\rangle + C\|y\|^p \label{eqn:banach}
\end{align}
Notice that if $B$ is a Hilbert space, we can take $p=2$ and $C=1$.

Given any sequence of random vectors $X_1,\dots,X_T$ in $B$, consider the sequence of real numbers $s_1,\dots,s_T$ defined recursively by:
\begin{enumerate}
    \item $s_0=0$
    \item If $\sum_{i=1}^{t-1} X_i\ne 0$, then we set: 
    \begin{align*}
        s_t = \sign\left(\sum_{i=1}^{t-1}s_i\right)\frac{\langle \nabla \|\sum_{i=1}^{t-1}X_i\|^p, X_t\rangle }{p\|\sum_{i=1}^{t-1}X_i\|^{p-1}}~,
    \end{align*} where $\sign(x)=1$ if $x\ge 0$, $-1$ if $x<0$ and $0$ if $x=0$.
    \item If $\sum_{i=1}^{t-1} X_i= 0$, set $s_t=0$.
\end{enumerate}

The critical property of this sequence is the following, proved in \cite{cutkosky2018algorithms} Theorem 5.3. We reproduce the proof for completeness below as we have mildly tightened the constants:
\begin{restatable}{Lemma}{thms}\label{thm:s}
Suppose $B$, $s_t$ and $X_t$ are as described above and $p\ge 1$. Then $|s_t|\le \|X_t\|$ for all $t$, and
\begin{align*}
    \left\|\sum_{t=1}^T X_t\right\|&\le \left|\sum_{t=1}^T s_t\right|+ \left(\max_{t\le T} \|X_t\|^p + C\sum_{t=1}^T \|X_t\|^p\right)^{1/p}
\end{align*}
\end{restatable}
\begin{proof}
For the bound on $s_t$, observe that since $x\mapsto \|x\|$ is by definition 1-Lipschitz, $\|(\nabla \|x\|)\|_\star\le 1$. Therefore we have 
\begin{align*}
    |s_t|&\le \frac{|\langle \nabla \|\sum_{i=1}^{t-1}X_i\|^p, X_t\rangle|}{p\|\sum_{i=1}^{t-1}X_i\|^{p-1}}\\
    &\le \langle \nabla \|\sum_{i=1}^{t-1}X_i\|, X_t\rangle\\
    &\le \|X_t\|
\end{align*}
Now, for the second statement we proceed by induction. Clearly the statement holds for $T=1$. Suppose now that
\begin{align*}
    \left|\sum_{t=1}^{T-1} X_t\right|&\le \left\|\sum_{t=1}^{T-1} s_t\right\| + \left(\max_{t\le T-1} \|X_t\|^p + C\sum_{t=1}^{T-1} \|X_t\|^p\right)^{1/p}
\end{align*}
Now, first observe that if $\sum_{i=1}^{T-1}X_i=0$, then $\|\sum_{i=1}^T X_i\|=\|X_i\|\le \max_{t\le T}\|X_i\|$ and so the statement holds. Let us now assume $\sum_{i=1}^{T-1}X_i\ne 0$
Define $X_{1:t} = \sum_{i=1}^t X_i$. Then we have:
\begin{align*}
    \left\|\sum_{i=1}^T X_i\right\|&=\left(\left\|X_{1:T-1} + X_T\right\|^p\right)^{1/p}\le \left(\|X_{1:T-1}\|^p + \langle \nabla \|X_{1:T-1}\|^p, X_T\rangle +  C\|X_T\|^p\right)^{1/p}
\end{align*}
Now we consider two cases, either $\|X_{1:T-1}\|^p + \langle \nabla \|X_{1:T-1}\|^p, X_T\rangle\le 0$ or not. If $\|X_{1:T-1}\|^p + \langle \nabla \|X_{1:T-1}\|^p, X_T\rangle\le 0$, then we have just shown:
\begin{align*}
    \left\|\sum_{i=1}^T X_i\right\|&\le C^{1/p}\|X_T\|\le \left|\sum_{i=1}^T s_i\right| + \left(\max_{t\le T} \|X_t\|^p + C\sum_{i=1}^T \|X_t\|^p\right)^{1/p}
\end{align*}
and so we are done. 

Instead, let us suppose $\|X_{1:T-1}\|^p + \langle \nabla \|X_{1:T-1}\|^p, X_T\rangle>0$. This implies $\|X_{1:T-1}\| + \frac{\langle \nabla \|X_{1:T-1}\|^p, X_T\rangle}{p\|X_{1:T-1}\|^{p-1}}\ge 0$ as well. Now, since the function $x\mapsto x^p$ is convex for any positive $x$, we have $x^p + p yx^{p-1}\le (x+y)^p$ for any $x\ge 0$ and $x+y\ge 0$. Thus:
\begin{align*}
    \|X_{1:T-1}\|^p + \langle \nabla \|X_{1:T-1}\|^p, X_T\rangle\le \left(\|X_{1:T-1}\| +\frac{\langle \nabla \|X_{1:T-1}\|^p, X_T\rangle}{p\|X_{1:T-1}\|^{p-1}}\right)^p
\end{align*}
Putting this together with our induction hypothesis:
\begin{align*}
    \left\|\sum_{i=1}^T X_i\right\|&\le \left(\left(\|X_{1:T-1}\| +\frac{\langle \nabla \|X_{1:T-1}\|^p, X_T\rangle}{p\|X_{1:T-1}\|^{p-1}}\right)^p +  C\|X_T\|^p\right)^{1/p}\\
    &\le \left(\left(\left|\sum_{i=1}^{T-1}s_i\right|+\left(\max_{i\le T-1}\|X_i\|^p + C\sum_{i=1}^{T-1}\|X_i\|^p\right)^{1/p}+\frac{\langle \nabla \|X_{1:T-1}\|^p, X_T\rangle}{p\|X_{1:T-1}\|^{p-1}}\right)^p  +  C\|X_T\|^p\right)^{1/p}\\
    &\le \left[\left(\left|\left|\sum_{i=1}^{T-1}s_i\right|+\frac{\langle \nabla \|X_{1:T-1}\|^p, X_T\rangle}{p\|X_{1:T-1}\|^{p-1}}\right|+\left(\max_{i\le T-1}\|X_i\|^p + C\sum_{i=1}^{T-1}\|X_i\|^p\right)^{1/p} \right)^p  +  C\|X_T\|^p\right]^{1/p}
\end{align*}
Now, following \cite{cutkosky2018algorithms}, we observe that for any positive $a$, $b$ and $c$, $(a+b)^p-b^p\le (a+b+c)^p-(b+c)^p$. Thus, setting $a=\left|\left|\sum_{i=1}^{T-1}s_i\right|+\frac{\langle \nabla \|X_{1:T-1}\|^p, X_T\rangle}{p\|X_{1:T-1}\|^{p-1}}\right|$, $b=\left(\max_{i\le T-1}\|X_i\|^p + C\sum_{i=1}^{T-1}\|X_i\|^p\right)^{1/p}$ and $b+c=\left(\max_{i\le T}\|X_i\|^p + C\sum_{i=1}^{T}\|X_i\|^p\right)^{1/p}$, we obtain:
\begin{align*}
    \left(\left|\left|\sum_{i=1}^{T-1}s_i\right|+\frac{\langle \nabla \|X_{1:T-1}\|^p, X_T\rangle}{p\|X_{1:T-1}\|^{p-1}}\right|+\left(\max_{i\le T-1}\|X_i\|^p + C\sum_{i=1}^{T-1}\|X_i\|^p\right)^{1/p} \right)^p&\le (a+b+c)^p +b^p - (b+c)^p\\
    &=(a+b+c)^p - C\|X_T\|^p
\end{align*}
Plugging this identity back into our previous bound on $\left\|\sum_{i=1}^T X_i\right\|$, we have:
\begin{align*}
    \left\|\sum_{i=1}^T X_i\right\|&\le \left|\left|\sum_{i=1}^{T-1}s_i\right|+\frac{\langle \nabla \|X_{1:T-1}\|^p, X_T\rangle}{p\|X_{1:T-1}\|^{p-1}}\right|+\left(\max_{i\le T}\|X_i\|^p + C\sum_{i=1}^{T}\|X_i\|^p\right)^{1/p}\\
    &=\left|\left|\sum_{i=1}^{T-1}s_i\right|+\sign\left(\sum_{i=1}^{t-1}s_i\right)s_T\right|+\left(\max_{i\le T}\|X_i\|^p +C \sum_{i=1}^{T}\|X_i\|^p\right)^{1/p}\\
    &=\left|\sum_{i=1}^{T}s_i\right|+\left(\max_{i\le T}\|X_i\|^p + C\sum_{i=1}^{T}\|X_i\|^p\right)^{1/p}
\end{align*}

\end{proof}

\subsection{Warm-up: Hilbert space}
Now, we show how to use Lemma \ref{thm:s} in conjuction with the standard 1-dimensional Freedman inequality to generate concentration bounds in Hilbert spaces. First, we state a (mildly weaker) version of the standard 1-dimensional inequality.

\begin{restatable}{Lemma}{thmFreedman}\label{thm:Freedman}
[Freedman's inequality] Suppose $D_1,D_2,\dots,D_T$ is a martingale difference sequence adapted to a filtration $F_1,F_2,\dots$ such that $D_i\le R$ almost surely for all $i$. Let $\E_i$ indicate expectation conditioned on $F_i$. Suppose further that for all $t$ with probability 1,
\begin{align*}
    \sigma_t^2 \ge \mathbb{E}_{t-1}[D_t^2]
\end{align*}
Then with probability at least $1-\delta$, for all $k$ we have
\begin{align*}
    \sum_{t=1}^k D_t\le \frac{2R\log(1/\delta)}{3} + \sqrt{2\sum_{t=1}^k\sigma_t^2\log(1/\delta)}
\end{align*}
\end{restatable}
\begin{proof}
From the standard inequality (see e.g. \cite{tropp2011freedman} Theorem 1.1), we have 
\begin{align*}
    P\left[\exists k\ :\ \sum_{t=1}^k D_t\ge \epsilon\right]\le \exp\left(-\frac{\epsilon^2/2}{\sum_{t=1}^k \sigma_t^2 + R\epsilon/3}\right)
\end{align*}
Now, set the RHS equal to $\delta$ to obtain:
\begin{align*}
    \frac{\epsilon^2/2}{\sum_{t=1}^k \sigma_t^2 + R\epsilon/3}=\log(1/\delta)
\end{align*}
The result now follows by using the quadratic formula to bound $\epsilon$.
\end{proof}

Now, we are in a position to describe our extension of Freedman's inequality to Hilbert spaces (which satisfy the hypotheses of Lemma \ref{thm:s} with $p=2$ and $C=1$):

\begin{restatable}{Lemma}{thmFreedmanbanach}\label{thm:Freedmanhilbert}
Suppose $X_1,\dots,X_T$ is a martingale difference sequence in a Hilbert space such that $\|X_t\|\le R$ almost surely for some constant $R$.
Further, assume $\E_{t-1}[\|X_t\|^2]\le \sigma_t^2$ with probability 1 for some constants $\sigma_t$.
Then with probability at least $1-3\delta$, for all $k$ we have:
\begin{align*}
    \left\|\sum_{t=1}^k X_t\right\|&\le 3R\max(1,\log(1/\delta)) \\
    &\quad+3\sqrt{\sum_{t=1}^k \sigma_t^2\max(1,\log(1/\delta))}
\end{align*}
\end{restatable}

\begin{proof}
Define $s_t$ as in Lemma \ref{thm:s}. Then with probability 1 we have for all $k$:
\begin{align*}
    \left\|\sum_{t=1}^kX_t\right\|\le \left|\sum_{t=1}^k s_t\right| + \sqrt{R^2+\sum_{t=1}^K \|X_t\|^2}
\end{align*}
Further, notice that $s_t$ is itself a martingale difference sequence, and satisfies $|s_t|\le \|X_t\|\le R$. Therefore by Lemma \ref{thm:Freedman}, with probability at least $1-2\delta$
\begin{align*}
    \left|\sum_{t=1}^k s_t\right|\le \frac{2R\log(1/\delta)}{3} + \sqrt{2\sum_{t=1}^k\sigma_t^2\log(1/\delta)}
\end{align*}
Now, for the second term, set $Z_t = \|X_t\|^2 -\E_{t-1}[\|X_t\|^2]$. Then $Z_t$ is also a martingale difference sequence. Since $\|X_t\|^2\le R^2$ with probability 1, $Z_t\le R^2$ with probability 1. Further, 
\begin{align*}
    \mathbb{E}_{t-1}[\|Z_t\|^2]\le \mathbb{E}_{t-1}[\|X_t\|^{4}]\le \mathbb{E}_{t-1}[\|X_t\|^2] R^2\le \sigma_t^2 R^2
\end{align*}
Thus, again by Lemma \ref{thm:Freedman}, with probability at least $1-\delta$:
\begin{align*}
    \sum_{t=1}^k Z_t &\le\frac{2R^2\log(1/\delta)}{3} +\sqrt{2\sum_{t=1}^T R^2\sigma_t^2\log(1/\delta)}
\end{align*}
which implies:
\begin{align*}
    &\sum_{t=1}^k\| X_t\|^2 \le\sum_{t=1}^k\sigma_t^2 + \frac{2R^2\log(1/\delta)}{3} + \sqrt{2\sum_{t=1}^k R^2\sigma_t^2\log(1/\delta)}
    \intertext{Applying Young inequality:}
    &\le \sum_{t=1}^k \sigma_t^2 + \frac{2R^2\log(1/\delta)}{3} + \sqrt{R^{4}\log^2(1/\delta)+\left(\sum_{t=1}^k\sigma_t^2\right)^2}\\
    &\le 2\sum_{t=1}^k \sigma_t^2 + \frac{5R^2\log(1/\delta)}{3}
\end{align*}
Putting everything together, with probability at least $1-3\delta$:
\begin{align*}
    &\left\|\sum_{t=1}^kX_t\right\|\le \left|\sum_{t=1}^k s_t\right| + \sqrt{R^2+\sum_{t=1}^K\|X_t\|^2}\\
    &\le  \frac{2R(\log(1/\delta)+3/2)}{3} + \sqrt{2\sum_{t=1}^k \sigma_t^2\log(1/\delta)}+\sqrt{2\sum_{t=1}^k \sigma_t^2 + \frac{5R^2\log(1/\delta)}{3}}\\
    &\le  3R\max(1,\log(1/\delta)) +3\sqrt{\sum_{t=1}^k \sigma_t^2\max(1,\log(1/\delta))}
\end{align*}
\end{proof}

\subsection{Extension to Banach space}
Having provided the concentration results in the more familiar Hilbert space setting, now we move to Banach spaces. We will also need the following useful observation:
\begin{Lemma}\label{thm:ptosqrt}
For any $0<p\le q$ and any positive $x_1,\dots,x_T$,
\begin{align*}
    \left(\sum_{t=1}^T x_t^q\right)^{1/q}\le \left(\sum_{t=1}^T x_t^p\right)^{1/p} 
\end{align*}
\end{Lemma}
\begin{proof}
We differentiate the expression $\left(\sum_{t=1}^T x_t^p\right)^{1/p}$ with respect to $p$ and show that the derivative is always negative, which suffices to prove the Lemma.

\begin{align*}
    \frac{d}{dp}\left(\sum_{t=1}^T x_t^p\right)^{1/p} &=\frac{d}{dp}\exp\left\{\frac{1}{p}\log\left[\sum_{t=1}^T \exp\left(p\log(x_t)\right)\right]\right\}\\
    &=\left(\sum_{t=1}^T x_t^p\right)^{1/p}\left(\frac{-\log\left[\sum_{t=1}^T \exp\left(p\log(x_t)\right)\right]}{p^2} + \frac{\sum_{t=1}^T \exp\left(p\log(x_t)\right)\log(x_t)}{p\sum_{t=1}^T\exp\left(p\log(x_t)\right)}\right)\\
    &=\left(\sum_{t=1}^T x_t^p\right)^{1/p}\left(\frac{-\log\left[\sum_{t=1}^T x_t^p\right]}{p^2} + \frac{ \sum_{t=1}^T x_t^p\log(x_t)}{p\sum_{t=1}^T x_t^p}\right)\\
    &=\frac{\left(\sum_{t=1}^T x_t^p\right)^{1/p}}{p^2}\left( -\log\left[\sum_{t=1}^T x_t^p\right] + \frac{ \sum_{t=1}^T x_t^p\log(x_t^p)}{\sum_{t=1}^T x_t^p}\right)\\
    &=\frac{\left(\sum_{t=1}^T x_t^p\right)^{1/p-1}}{p^2}\left( \sum_{t=1}^T x_t^p\log(x_t^p) -\log\left[\sum_{t=1}^T x_t^p\right]\left(\sum_{t=1}^T x_t^p\right) \right)
    \intertext{since $\log$ is an increasing function and $x_t> 0$:}
    &\le \frac{\left(\sum_{t=1}^T x_t^p\right)^{1/p-1}}{p^2}\left( \sum_{t=1}^T x_t^p\log\left[\sum_{t=1}^T x_t^p\right] -\log\left[\sum_{t=1}^T x_t^p\right]\left(\sum_{t=1}^T x_t^p\right) \right)\\
    &=0
\end{align*}
\end{proof}

\begin{restatable}{Lemma}{thmFreedmanbanach2}\label{thm:Freedmanbanach}
Suppose $X_1,\dots,X_T$ is a martingale difference sequence in a Banach space satisfying (\ref{eqn:banach}) for $p\in(1,2]$ such that $\|X_t\|\le R$ almost surely for some constant $R$.
Further, assume $\E_{t-1}[\|X_t\|^2]\le \sigma_t^2$ with probability 1 for some constants $\sigma_t$.
Then with probability at least $1-3\delta$, for all $k$ we have:
\begin{align*}
    \left\|\sum_{t=1}^k X_t\right\|&\le \frac{2R(\log(1/\delta)+3/2)}{3} + \sqrt{2\sum_{t=1}^k \sigma_t^2\log(1/\delta)}+\left(2C\sum_{t=1}^k \sigma_t^p + \frac{7CR^p\log(1/\delta)}{3}\right)^{1/p}
\end{align*}
Additionally, if $C\ge 1$ and $p\le 2$:
\begin{align*}
    5CR\max(1,\log(3/\delta))+ 4\left(C\sum_{t=1}^k \sigma_t^p\right)^{1/p}\sqrt{\max(1, \log(3/\delta))}
\end{align*}
\end{restatable}

\begin{proof}
First, note that by Jensen, we have $\E_{t-1}[\|X_t\|^p]\le \E_{t-1}[\|X_t\|^2]^{p/2}\le \sigma_t^p$.

Define $s_t$ as in Lemma \ref{thm:s}. Then with probability 1 we have for all $k$:
\begin{align*}
    \left\|\sum_{t=1}^kX_t\right\|\le \left|\sum_{t=1}^k s_t\right| + \left(R^p+C \sum_{t=1}^K \|X_t\|^p\right)^{1/p}
\end{align*}
Further, notice that $s_t$ is itself a martingale difference sequence, and satisfies $|s_t|\le \|X_t\|\le R$. Therefore by Lemma \ref{thm:Freedman}, with probability at least $1-2\delta$
\begin{align*}
    \left|\sum_{t=1}^k s_t\right|\le \frac{2R\log(1/\delta)}{3} + \sqrt{2\sum_{t=1}^k\sigma_t^2\log(1/\delta)}
\end{align*}
Now, for the second term, set $Z_t = \|X_t\|^p -\E_{t-1}[\|X_t\|^p]$. Then $Z_t$ is also a martingale difference sequence. Since $\|X_t\|^p\le R^p$ with probability 1, $Z_t\le 2R^p$ with probability 1. Further, 
\begin{align*}
    \E_{t-1}[\|Z_t\|^2]\le \E_{t-1}[\|X_t\|^{2p}]\le \E_{t-1}[\|X_t\|^p] R^p\le \sigma_t^p R^p
\end{align*}
Thus, again by Lemma \ref{thm:Freedman}, with probability at least $1-\delta$:
\begin{align*}
    \sum_{t=1}^k Z_t &\le\frac{4R^p\log(1/\delta)}{3} +\sqrt{2\sum_{t=1}^T R^p\sigma_t^p\log(1/\delta)}
\end{align*}
\begin{align*}
    \sum_{t=1}^k\| X_t\|^p &\le\sum_{t=1}^k\sigma_t^p + \frac{4R^p\log(1/\delta)}{3} + \sqrt{2\sum_{t=1}^k R^p\sigma_t^p\log(1/\delta)}
    \intertext{Applying Young inequality:}
    &\le \sum_{t=1}^k \sigma_t^p + \frac{4R^p\log(1/\delta)}{3} + \sqrt{R^{2p}\log^2(1/\delta)+\left(\sum_{t=1}^k\sigma_t^p\right)^2}\\
    &\le 2\sum_{t=1}^k \sigma_t^p + \frac{7R^p\log(1/\delta)}{3}
\end{align*}
Putting everything together, with probability at least $1-3\delta$:
\begin{align*}
    &\left\|\sum_{t=1}^kX_t\right\|\le \left|\sum_{t=1}^k s_t\right| + \left(R^p+C\sum_{t=1}^K\|X_t\|^p\right)^{1/p}\\
    &\le  \frac{2R(\log(1/\delta)+3/2)}{3} + \sqrt{2\sum_{t=1}^k \sigma_t^2\log(1/\delta)}+\left(2C\sum_{t=1}^k \sigma_t^p + \frac{7CR^p\log(1/\delta)}{3}\right)^{1/p}
\end{align*}
Now, if $C\ge 1$, we can apply Lemma \ref{thm:ptosqrt} and over approximate to obtain with probability $1-3\delta$:

\begin{align*}
    \left\|\sum_{t=1}^kX_t\right\|&\le  \frac{2R(\log(1/\delta)+3/2)}{3} + \left(\sum_{t=1}^k \sigma_t^p\right)^{1/p}\sqrt{2\log(1/\delta)}+\left(2C\sum_{t=1}^k \sigma_t^p\right)^{1/p} + \left(\frac{7CR^p\log(1/\delta)}{3}\right)^{1/p}\\
    &\le 5CR\max(1,\log(1/\delta))+ 4\left(C\sum_{t=1}^k \sigma_t^p\right)^{1/p}\sqrt{\max(1, \log(1/\delta))}
\end{align*}

\end{proof}
\begin{figure*}[h!]
\centering
  \begin{subfigure}
     \centering
     \includegraphics[width=0.45\textwidth]{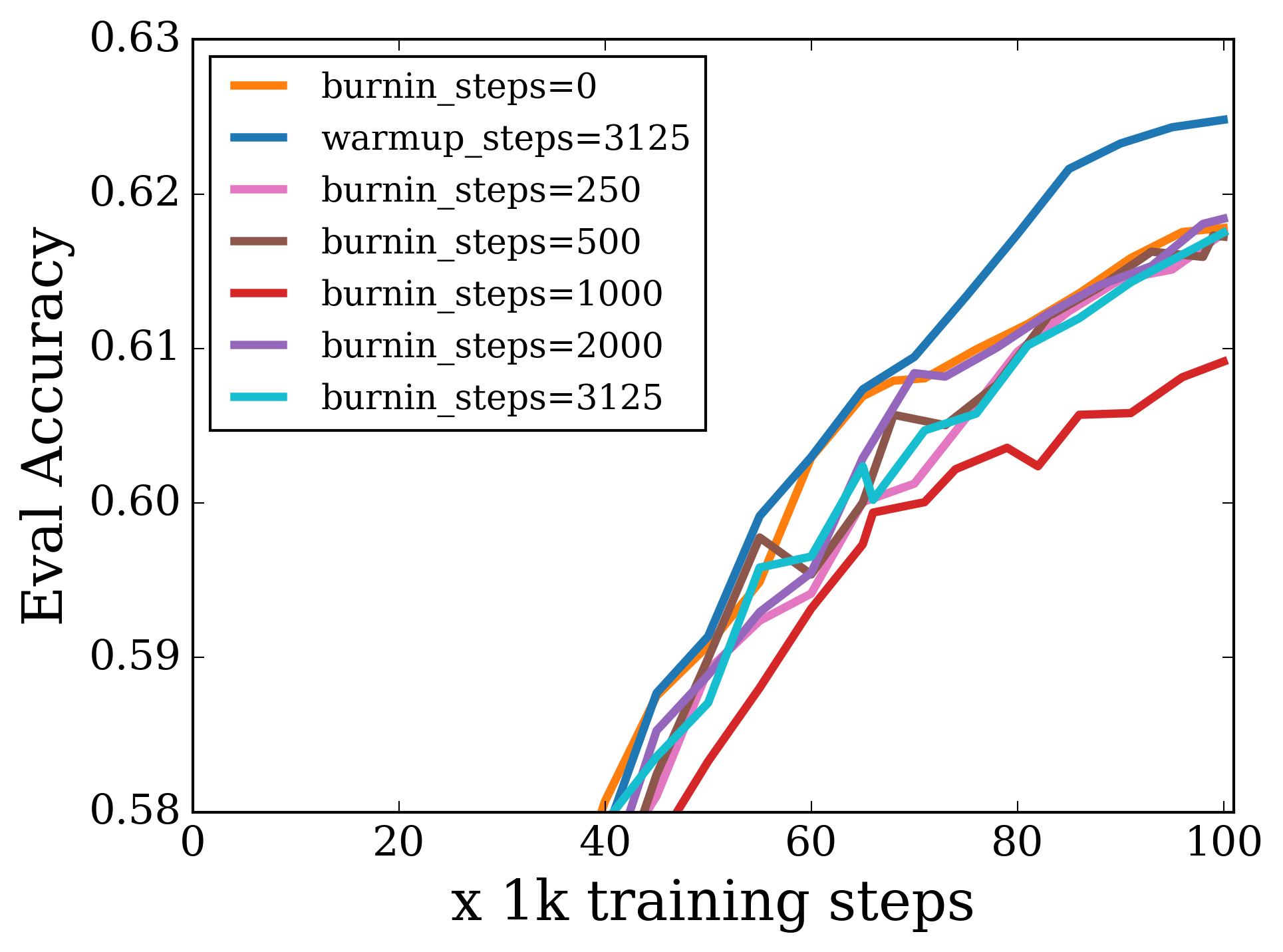}
     \label{fig:bert_base_alg1}
    \qquad
  \end{subfigure}
    \begin{subfigure}
     \centering
     \includegraphics[width=0.45\textwidth]{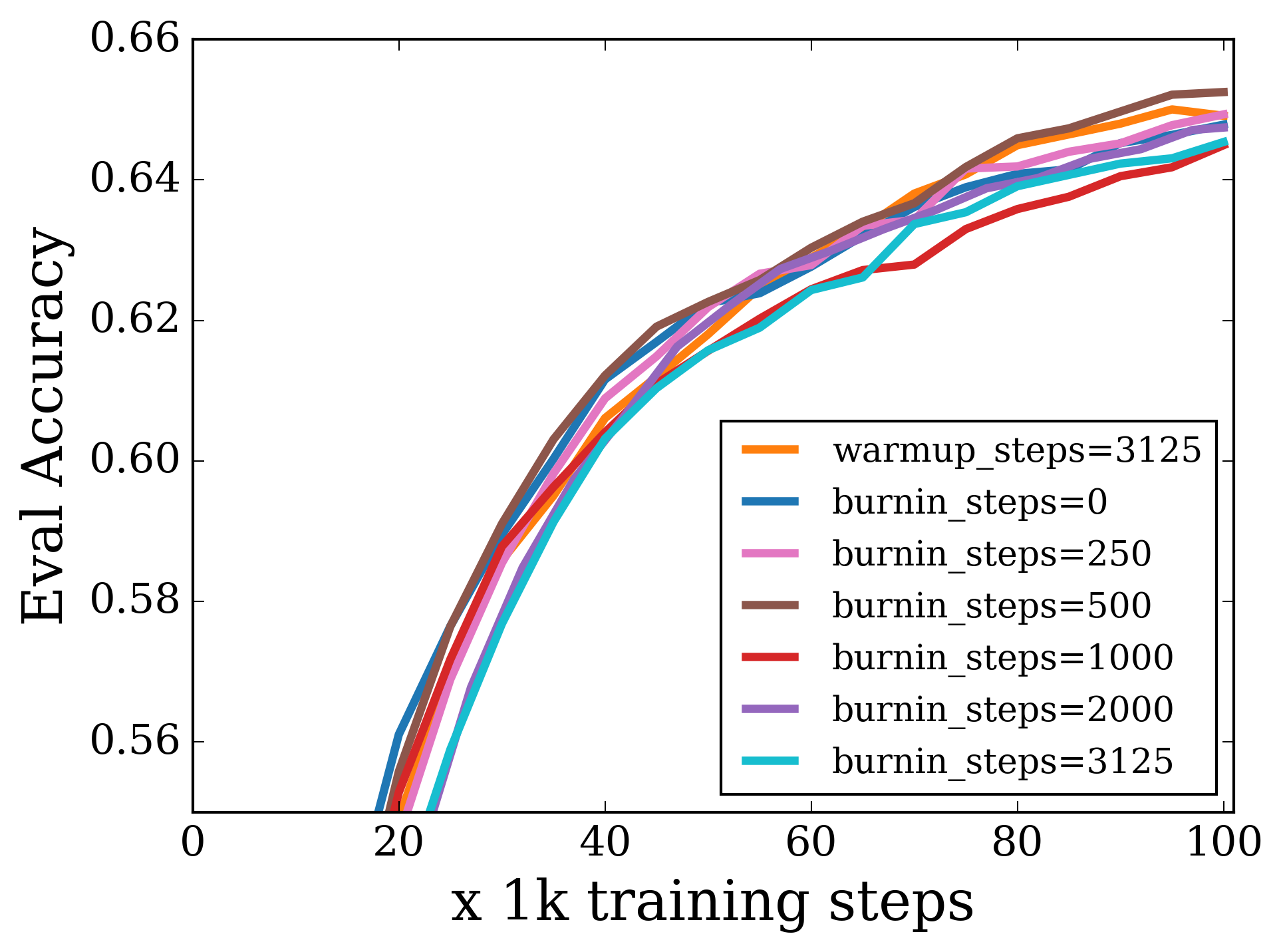}
     \label{fig:bert_base}
  \end{subfigure}
\caption{Comparison between warm-up and burn-in strategies on Masked language modeling BERT-Base pre-training task. Plot (a) shows the eval accuracy when trained with Algorithm \ref{alg:nsgdclip} and plot (b) does the same when trained with LAMB.}
  \label{fig:experiments_appendix}
\end{figure*}
\section{Experiments}\label{sec:experiments_appendix}
In this section, we run additional experiments on BERT-Base pre-training task. BERT-Base is a smaller version of BERT-Large model we experimented in Section \ref{sec:experiments}. Again, all our experiments were conducted using the Tensorflow framework on TPUv3 architecture. When using Algorithm \ref{alg:nsgdclip}, we train all the models shown using the base learning rate $\eta_{0}$ of 0.5 and batch size of 512. We came up with this learning rate by performing grid search $\eta_{0} \in [1.0, 0.5, 0.1, 0.05, 0.01, 0.005, 0.001, 0.0005, 0.0001]$ and choosing the one which attains the best eval accuracy. Again, shown in Fig \ref{fig:experiments_appendix}, our warmup baseline employs the standard practice of linear warm-up for 3125 steps and polynomical decay of $\eta_{t} = \eta_{0}*(1 - \frac{t}{T})$ for the rest of the steps. For a fair comparison with the baseline, we start the polynomial decay after 3125 steps regardless of how many steps were used for burn-in. We set the learning rate to the constant base learning rate value of 0.5 for for steps in between burn-in and polynomial decay. For the models trained using burn-in, instead of linear warm-up, we set the base learning rate $\eta_{0}$ to zero for the amount of steps we run the burn-in for. As shown in Fig \ref{fig:experiments_appendix}, we perform a grid search across $[0, 250, 500, 1000, 2000, 3125]$ steps for the initial burn-in period and compare their results. When training the smaller BERT-Base model we observe that burn-in is less crucial compared to when training BERT-Large i.e. the model converges regardless of the amount of burn-in employed. Moreover, we don't see any convergence issues when we don't employ even any warm-up, this suggests that warm-up is also much less crucial (similar to burn-in) on the simpler BERT-Base problem.

\section{Smoothness Lemmas in Banach Spaces}\label{sec:smoothbanach}
In this section we provide a few technical lemmas relating to smoothness in Banach spaces. Most of these are fairly straightforward generalizations of standard results in Hilbert spaces.

We begin with a notations and definitions. Recall that a \emph{Banach space} over the reals is a real vector space $\B$ together with a norm $\|\cdot\|:\B\to \R$ such that $\B$ is complete with respect to the topology induced by the norm. The dual space $\B^\star$ is the space of bounded linear functions $\B\to \R$. When $\B$ is a Hilbert space, there is a natural isomorphism between $i:\B^\star\to \B$ such that for any $v\in \B^\star$ and $w\in \B$, $v(w)=\langle i(v),w\rangle$. In order to preserve notational familiarity, even when $\B$ is \emph{not} a Hilbert space, we therefore adopt the notation $\langle v,w\rangle$ to refer to $v(w)$, the application of the linear function $v\in\B^\star$ to the vector $w\in \B$. Note that in this way the ordering of the arguments to $\langle \cdot,\cdot\rangle$ is important: $\langle w, v\rangle$ is meaningless. The dual space $\B^\star$ is equipped with the \emph{dual norm} $\|v\|_\star =\sup_{\|x\|\le 1}\langle v,x\rangle$, and is itself a Banach space. 

A function $F:\B\to \R$ is Frechet differentiable if for all $w\in \B$ there is a bounded linear operator $v\in \B^\star$ such that:
\begin{align*}
    \lim_{\delta\to 0} \frac{|F(w+\delta) - F(w) - \langle v,\delta\rangle|}{\|\delta\|} =0
\end{align*}
the operator $v$ is denoted $\nabla F(w)$. Note that if $F$ is Frechet differentiable, it also satisfies the familiar directional derivative identity: if $f(t) = F(w+t\delta)$, then
\begin{align*}
    f'(t) = \langle \nabla F(w+t\delta), \delta\rangle 
\end{align*}
 
The second derivative or Hessian of $F$ at a point $w$ is a linear map $\nabla^2F(w):\B\to \B^\star$ satisfying:
\begin{align*}
    \lim_{\delta\to 0} \frac{\|\nabla F(w+\delta) -\nabla F(w) - \nabla^2F(w) \delta\|_\star}{\|\delta\|}=0   
\end{align*}

In this paper we exclusively consider spaces such that $\B^\star$ satisfies:
\begin{align}
    \|x+y\|_\star^p\le \|x\|_\star^p+\langle \nabla \|x\|_\star^p,y\rangle + C\|y\|_\star^p
\end{align}
for some $p$ and $C$ for all $x,y\in\B^\star$. We will also assume that $C\ge 1$.

\begin{restatable}{Proposition}{propsmooth}\label{prop:smooth}
Suppose that $F:\B\to \R$ is Frechet differentiable and the gradient satisfies $\|\nabla F(x)-\nabla F(y)\|_\star \le L\|x-y\|$ for all $x,y\in \B$ (i.e. $F$ is $L$-smooth). Then for all $x,y\in \B$:
\begin{align*}
    F(x)\le F(y)  + \langle \nabla F(y),x-y\rangle + \frac{L}{2}\|x-y\|^2
\end{align*}
\end{restatable}
\begin{proof}
Consider the function $g:\R\to \R$ given by $g(t) = F(y+t(x-y))$. Since $F$ is Frechet differentiable, $g$ is differentiable and $g'(t) = \langle \nabla F(y+t(x-y)),(x-y)\rangle$. Therefore by the fundamental theorem of calculus:
\begin{align*}
    g(1)&=g(0) + \int_0^1 g'(t) dt\\
    &=g(0) + g'(0) + \int_0^1(g'(t)-g'(0))dt\\
    &\le g(0) + \langle \nabla F(y),x-y\rangle + \int_0^1 |g'(t)-g'(0)|dt\\
    &\le g(0) + \langle \nabla F(y),x-y\rangle  +\int_0^1 |\langle \nabla F(y+t(x-y)) - \nabla F(y),x-y\rangle| dt\\
    &\le g(0) + \langle \nabla F(y),x-y\rangle  + \int_0^1 tL\|x-y\|^2 dt\\
    &=g(0) + \langle \nabla F(y),x-y\rangle  + \frac{L}{2}\|x-y\|^2|
\end{align*}
Now plug in the definition of $g(1)$ and $g(0)$ to conclude the argument.
\end{proof}

\begin{restatable}{Proposition}{propsecondsmooth}\label{prop:secondsmooth}
Suppose that $F:\B\to \R$ is twice Frechet differentiable and the Hessian satisfies $\|(\nabla^2 F(x)-\nabla^2 F(y))v\|_\star \le \rho \|v\|\|x-y\|$ for all $x,y,v\in \B$ (i.e. $F$ is $\rho$-second-order smooth). Then for all $x,y\in \B$:
\begin{align*}
    F(x)\le F(y)  + \langle \nabla F(y),x-y\rangle + \frac{\langle \nabla^2 F(y)(x-y),x-y\rangle}{2} + \frac{\rho}{6}\|x-y\|^3
\end{align*}
\end{restatable}
\begin{proof}
Consider the function $g:\R\to \R$ given by $g(t) = F(y+t(x-y))$. Since $F$ is twice Frechet differentiable, $g$ is twice differentiable and $g'(t) = \langle \nabla F(y+t(x-y)),x-y\rangle$ and $g''(t) = \langle \nabla^2F(y+t(x-y))(x-y),x-y\rangle$. Therefore by the fundamental theorem of calculus:
\begin{align*}
    g(1)&=g(0) + \int_0^1 g'(t) dt\\
    &=g(0) + \int_0^1 \left(g'(0) + \int_0^t g''(k)dk\right) dt\\
    &=g(0) + g'(0) + \int_0^1\int_0^t g''(k)dk dt\\
    &=g(0) + g'(0) + \int_0^1\int_0^t g''(0)dk dt+ \int_0^1\int_0^t (g''(k)- g''(0))dk dt\\
    &=g(0) + g'(0) + \frac{g''(0)}{2} + \int_0^1\int_0^t (g''(k)- g''(0))dk dt\\
    &= g(0) + g'(0) + \frac{g''(0)}{2} + \int_0^1\int_0^t \langle (\nabla^2 F(y+k(x-y)) -\nabla^2 F(y))(x-y),x-y\rangle dkdt\\
    &\le  g(0) + g'(0) + \frac{g''(0)}{2} + \int_0^1\int_0^t \rho \|x-y\|^3 k dkdt\\
    &\le g(0) + g'(0) + \frac{g''(0)}{2} + \frac{\rho \|x-y\|^3}{6}
\end{align*}
Now plug in the definition of $g(1)$ and $g(0)$ to conclude the argument.
\end{proof}

\begin{restatable}{Proposition}{propsecondsmoothgrad}\label{prop:secondsmoothgrad}
Suppose that $F:\B\to \R$ is twice Frechet differentiable and the Hessian satisfies $\|(\nabla^2 F(x)-\nabla^2 F(y))v\|_\star \le \rho \|v\|\|x-y\|$ for all $x,y,v\in \B$ (i.e. $F$ is $\rho$-second-order smooth). Then for all $x,y\in \B$:
\begin{align*}
    \|\nabla F(y) -  F(x)  - \nabla^2 F(x)(y-x)\|_\star \le \frac{\rho}{2}\|x-y\|^2
\end{align*}
\end{restatable}
\begin{proof}
For any fixed $z\in \B$, define $g_z(t) = \langle \nabla F(x + t(y-x)),z\rangle$. Then we have $g_z'(t) = \langle \nabla^2 F(x + t(y-x))(y-x),z\rangle$. Then, by the fundamental theorem of calculus:
\begin{align*}
    g_z(1) &= g_z(0)+\int_0^1 g_z'(t) dt\\
    &\le g_z(0) + g_z'(0) + \int_0^1 |g_z'(t)-g_z'(0)|dt\\
    &\le  g_z(0) + g_z'(0) + \int_0^1 \rho t\|x-y\|^2 \|z\|dt\\
    &\le g_z(0) +g_z'(0) + \frac{\rho \|x-y\|^2 \|z\|}{2}
    \intertext{rearrange and apply the definition of $g_z$:}
    \langle \nabla F(y) - \nabla F(x) - \nabla^2 F(x)(y-x),z\rangle&\le \frac{\rho \|x-y\|^2 \|z\|}{2}
\end{align*}
Since $z$ was arbitrary, by definition of dual norm we have
\begin{align*}
    \|\nabla F(y) - \nabla F(x) - \nabla^2 F(x)(y-x)\|_\star \le \frac{\rho}{2}\|x-y\|^2
\end{align*}

\end{proof}

The following important Lemma generalizes Lemma 2 of \cite{cutkosky2020momentum} to more general norms, and improves constants.
\lemonestep*
\begin{proof}
First, by smoothness (Proposition \ref{prop:smooth}) we have
\begin{align*}
    F(w') \le F(w) + \langle \nabla F(w), w'-w\rangle + \frac{L}{2}\|w'-w\|^2
\end{align*}

Next, since $\|g\|=1$,
\begin{align*}
    F(w')&\le F(w) - \eta \langle \nabla F(w), g\rangle + \frac{L\eta^2}{2}\\
    &= F(w) - \eta \langle g^\star -\epsilon, g\rangle + \frac{L\eta^2}{2}\\
    &=F(w) - \eta (\|g^\star\|_\star - \langle \epsilon, g\rangle)+ \frac{L\eta^2}{2}\\
    &\le F(w) - \eta \|\nabla F(w) + \epsilon\|_\star + \eta\|\epsilon\|_\star \|g\|+ \frac{L\eta^2}{2}\\
    &\le F(w) - \eta \|\nabla F(w)\|_\star + 2\eta \|\epsilon\|_\star + \frac{L\eta^2}{2}
\end{align*}
\end{proof}

\end{document}